\documentclass{article}

\usepackage[final,nonatbib]{neurips_2026}
\usepackage[numbers]{natbib}

\usepackage[T1]{fontenc}
\usepackage{microtype}
\usepackage{xcolor}
\usepackage{nicefrac}

\makeatletter
\newcommand{\ssymbol}[1]{\@fnsymbol{#1}}
\makeatother

\usepackage{microtype}
\usepackage{graphicx}
\usepackage{subcaption}
\usepackage{booktabs}
\usepackage{multirow}
\usepackage{tabularx}
\usepackage[hidelinks]{hyperref}
\usepackage{amsmath,amssymb,amsthm}
\usepackage[capitalize,noabbrev]{cleveref}
\usepackage{xcolor}
\definecolor{accentgreen}{HTML}{3FB950}
\definecolor{accentorange}{HTML}{F0883E}

\theoremstyle{plain}
\newtheorem{theorem}{Theorem}[section]
\newtheorem{proposition}[theorem]{Proposition}

\theoremstyle{definition}

\theoremstyle{remark}
\newtheorem{remark}[theorem]{Remark}

\graphicspath{{./}{./figures/}}

\title{Gluing Local Contexts into Global Meaning:\\ A Sheaf-Theoretic Decomposition of Transformer Representations}

\author{%
  Bryce Grant \\
  Case Western Reserve University \\
  \texttt{bag100@case.edu} \\
  \And
  Peng Wang \\
  Case Western Reserve University \\
  \texttt{pxw335@case.edu} \\
}

\begin{document}
\maketitle

\begin{abstract}
Prior research into interpretability does not test whether a representation stays consistent under paraphrasing. Each pair of paraphrases can be thought of as a local witness that two activations should be the same. From these local witnesses we can build a cellular sheaf on the paraphrase graph. The question of what is consistent under paraphrasing can be answered by the Laplacian $L_\mathcal{F}$ given a frozen model. No training required. Our kernel $H^0$ gives us a content subspace. We show that with only 20 dims the kernel outperforms the subspace that LEACE retains with all dims on held-out CounterFact fact retrieval by $+17.1$\,pp on average across 8 models. By constraining all paraphrases of a fact together we see an additional increase of $+6.5$\,pp. Our spectral complement shows us where the causal influence is concentrated. Ablating the coordinates that vary the most when paraphrased (which approximates ablating the spectral complement at the coordinate level) leads to $5.6$ to $26.5\times$ larger changes in model outputs compared to ablating coordinates with matched per-coordinate variability. We find this across 9 models from 124M to 13B params. By stacking layers we can create a grid which has $b_1 = N(L-1)$ algebraic 1-cycles. The harmonic mass derived from the model varies by nearly 3 orders of magnitude between the 6 architectures. It also tracks the order of steering fragility (Spearman's $\rho = 0.90$, exact one-sided permutation $p = 0.014$). The kernel, its spectral complement and the harmonic mass all come from the same paraphrase graph. Project page: \url{https://cwru-aism.github.io/gluing-lc-page/}.
\end{abstract}

\section{Introduction}
\label{sec:intro}

Transformers entangle what a sentence means with how it is written. ``The Eiffel Tower is in Paris'' and ``Paris is home to the Eiffel Tower'' have different representations even though they mean the same thing. There are techniques that decompose a representation into features \cite{bricken2023monosemanticity,cunningham2024sparse}, see if pre-specified properties are linearly decodable \cite{belinkov2022probing}, or compare two representation spaces \cite{kornblith2019similarity,raghu2017svcca}. However none of them deal with consistency between different contexts. That is, if we rephrase a sentence so it keeps the same meaning, will a feature still carry the same information? This becomes an issue when trying to steer models. Some methods may work on certain models and fail or even degrade performance on others \cite{dasilva2025steering}. Even within one model, steerability can vary widely with the prompt \cite{tan2025analysing}. There is currently no tool to know what regime your model falls under. We frame this problem as one of local-to-global consistency. Each pair of paraphrases is a local witness, evidence that the corresponding 2 activations should agree, and all together they would be expected to give a global consensus.

We create a cellular sheaf over the graph of paraphrases. Here we assign hidden states to the nodes and connect paraphrases with edges. Then we learn a restriction map $P^\top$ ($P \in \mathbb{R}^{d \times k}$) that maps each pair to a common space of dimension $k$. The coboundary $\delta^0$ gives us the disagreement on each edge. The sheaf Laplacian $L_\mathcal{F} = (\delta^0)^\top \delta^0$ has a kernel $H^0 = \ker(\delta^0)$ of globally consistent sections and a spectral complement of high-disagreement directions. We will refer to these directions as the ``$H^1$ directions'' (Remark~\ref{rem:h1} explains the notation).

\begin{figure}[t]
\centering
\includegraphics[width=0.92\linewidth]{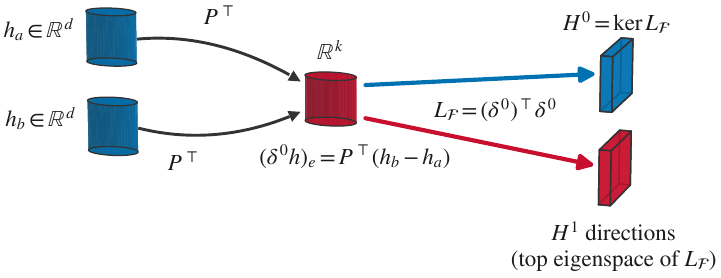}
\caption{Sheaf decomposition of a paraphrase pair. Hidden states $h_a, h_b$ pass through $P^\top$ into the edge stalk $\mathbb{R}^k$, where the coboundary measures their disagreement, $(\delta^0 h)_e = P^\top(h_b - h_a)$. The bottom and top eigenspaces of the sheaf Laplacian $L_\mathcal{F}$ split the edge stalk $\mathbb{R}^k$ into $H^0$ (blue, content stable across phrasing) and the $H^1$ directions (red, context-dependent variation).}
\label{fig:pipeline}
\end{figure}

If we only use one layer, the matching graph has no cycles, so algebraic $H^1 = 0$ and our construction is simply the projected within-pair covariance (Proposition~\ref{prop:laplacian_reduction}). If we stack the same activations over multiple layers we get cycles. It is natural to consider a grid of all (sentence, layer) combinations, with horizontal edges linking the paraphrases within each layer, and vertical edges linking the same sentence between adjacent layers. There is then one algebraic 1-cycle for every square in this grid, $b_1 = N(L-1)$ in total ($b_1$, the first Betti number, is the number of independent cycles). Each square asks whether passage from layer $l$ to layer $l{+}1$ commutes with paraphrase equivalence. The model-derived Hodge harmonic mass (the share of a model-derived edge cochain's energy in the harmonic space, supported on cycles but neither a gradient nor a curl) at $L=2$ ranges over nearly 3 orders of magnitude across 6 architectures ($10^{-4}$ to $0.08$; Table~\ref{tab:multilayer}) against a $0.25$ topological reference, and \S\ref{sec:fragility} tests it as a diagnostic of steering fragility. When we constrain all paraphrases of a fact together, $H^0$ outperforms pair-only $H^0$ by $+6.5$\,pp on all 8 architectures. A graph with the same number of edges and no cycle confirms that this improvement comes from constraining the paraphrases jointly, and the cycle itself does not change retrieval (\S\ref{sec:hypergraph_cohomology}).

Our contributions are:
{\setlength{\leftmargini}{1.4em}\setlength{\topsep}{2pt}\setlength{\partopsep}{0pt}
\begin{itemize}\setlength{\itemsep}{2pt}\setlength{\parskip}{0pt}
\item \textbf{Content from the kernel.} Sheaf $H^0$ at 20 dimensions beats LEACE's preserved subspace at full hidden dimension on held-out CounterFact retrieval across eight architectures (mean $+17.1$\,pp; $85.3\%$ vs.\ $43.1\%$ on GPT-2). Random or identity restriction maps keep most of the gain, and constraining all of a fact's paraphrases jointly adds $+6.5$\,pp (\S\ref{sec:content}).
\item \textbf{Influence from the spectral complement.} Ablating the most paraphrase-variable coordinates, a coordinate-wise proxy for the top eigenspace of $L_\mathcal{F}$, changes model outputs $5.6$ to $26.5\times$ more than ablating coordinates of matched per-coordinate variability on nine architectures from 124M to 13B parameters. Projecting out the sheaf $H^1$ eigenbasis shifts next-token distributions $17.8$--$398\times$ more than random subspaces and collapses fact recall to $0$--$5\%$, while matched random ablation leaves recall at baseline on the 7B models (\S\ref{sec:influence}, Appendices~\ref{app:sheaf_basis} and~\ref{app:illusion}).
\item \textbf{Fragility from harmonic mass.} Multi-layer grids carry $b_1 = N(L-1)$ algebraic 1-cycles and a model-derived harmonic mass spanning $[10^{-4}, 0.08]$ that ranks six architectures by steering fragility ($\rho = 0.90$, $p = 0.014$). On Llama-2-7B, contrastive sheaf steering preserves $31.0\%$ of facts against $4.2\%$ for random steering (\S\ref{sec:fragility}).
\end{itemize}
}

\section{Cellular Sheaves on Paraphrase Graphs}
\label{sec:background}

A cellular sheaf $\mathcal{F}$ on a graph $G = (V, E)$ assigns a vector space (stalk) $\mathcal{F}_v = \mathbb{R}^{d_v}$ to each node and a stalk $\mathcal{F}_e = \mathbb{R}^{d_e}$ to each edge, connected by linear restriction maps $\mathcal{F}_{v \trianglelefteq e}: \mathcal{F}_v \to \mathcal{F}_e$ at node-edge incidences \cite{curry2014sheaves,hansen2019toward}. The coboundary operator $\delta^0: C^0(\mathcal{F}) \to C^1(\mathcal{F})$ measures local disagreement:
\begin{equation}
(\delta^0 x)_e = \mathcal{F}_{v \trianglelefteq e}(x_v) - \mathcal{F}_{u \trianglelefteq e}(x_u)
\end{equation}
Here $e = (u, v)$ is an edge. If $\delta^0 x = 0$ for every edge, the local data is \emph{globally consistent} and combines to give a global section. The sheaf Laplacian $L_\mathcal{F} = (\delta^0)^\top \delta^0$ is a generalization of the usual graph Laplacian \cite{hansen2019toward}. Figure~\ref{fig:pipeline} summarizes the steps, starting with stalks at each node, applying $\delta^0$ and splitting into $H^0$ and $H^1$. As $x^\top L_\mathcal{F} x = \|\delta^0 x\|^2$, it follows that $\ker(L_\mathcal{F}) = \ker(\delta^0) = H^0(\mathcal{F})$ \cite{hansen2019toward}, the space of global sections. Failures to glue together local data are measured by the first cohomology group $H^1 = \ker(\delta^1) / \mathrm{im}(\delta^0)$. For a graph $\delta^1 = 0$, so $H^1 = C^1/\mathrm{im}(\delta^0)$: functions on edges which do not arise as the coboundary of a function on nodes.

Our interpretation of $H^0$ is that it contains the information that is preserved through paraphrase. We can think of the spectral complement of $L_\mathcal{F}$ as information that varies with context, such as syntax, word choice and surface form \cite{elhage2022toy,bricken2023monosemanticity}. By definition of the kernel, if all paraphrase pairs of a fact agree on a direction, it will stay constant when the fact is paraphrased, and so it tracks the fact's content. The kernel is made up of these directions. Cycles are useful for detecting where steering may fail. If the cochain computed from the model does not sum to zero around some cycle, that is, has a non-vanishing harmonic part, this means that the model's layer-to-layer maps do not commute with paraphrase equivalence. So information about a fact will be propagated inconsistently, and in general reading a direction off from one path will give a different result along another. A matching graph (one edge per paraphrase pair) does not have any cycles, so $b_1(G) = 0$, and if $P$ has full column rank $k$ then $\delta^0$ is surjective onto $C^1$ and we have algebraic $H^1 = \ker(\delta^1)/\mathrm{im}(\delta^0) = 0$.
For transformers we place hidden states on the nodes and edges correspond to paraphrase pairs. A restriction map $P^\top$ maps the full activation space to a $k$-dimensional edge space. The spectral decomposition is taken in edge space, and its eigenvectors are pulled back to the activation space using $P$. The choice of $P$ (PCA, canonical correlation analysis (CCA), or a contrastive Fisher discriminant) is the only knob, and it changes the decomposition.

LEACE (least-squares concept erasure)~\cite{belrose2023leace} learns a concept eraser, which is a good fit for removing the paraphrase / non-paraphrase distinction. For content stability we want a complementary target, so we look for the subspace with the smallest variation within paraphrase pairs. Sparse autoencoders \cite{bricken2023monosemanticity,cunningham2024sparse} learn features that make up one representation. Centered kernel alignment (CKA) and SVCCA \cite{kornblith2019similarity,raghu2017svcca} measure how similar two representation spaces are. It would be useful to decompose each representation space into context-independent and context-dependent components. Then we could attempt to manipulate only the context-dependent directions, leaving the underlying content unchanged. The coboundary provides such a decomposition, and the interventions in \S\ref{sec:experiments} use these subspaces.

\section{Local Construction on Matching Graphs}
\label{sec:methodology}

\textbf{Sheaf.} The interpretability sheaf $\mathcal{F}$ over a context graph $G = (V, E)$ assigns node stalks $\mathcal{F}(v) = \mathbb{R}^d$, in which the hidden states of one context form a 0-cochain, edge stalks $\mathcal{F}(e) = \mathbb{R}^k$ ($k \ll d$) as a shared comparison space, and linear restriction maps $\mathcal{F}_{v \trianglelefteq e} = P^\top$. The matching graph ($|V| = 2N$, $|E| = N$, no higher simplices) is the simplest case. We learn $P \in \mathbb{R}^{d \times k}$ via three concrete choices: joint PCA (top $k$ components of paired activations), CCA (solving $C_{12} C_{22}^{-1} C_{21} w = \lambda C_{11} w$ for the top $k$ eigenvectors), or contrastive Fisher. The contrastive Fisher map solves $C_W w = \lambda C_T w$ for $C_W = \tfrac{1}{N}\sum_i \Delta_i \Delta_i^\top$ and $C_T = \tfrac{1}{N}\sum_i h_i h_i^\top$, where $\Delta_i = h_i^{(1)} - h_i^{(2)}$ is the activation difference of pair $i$, and takes the $k$ \emph{largest} generalized eigenvectors as columns of $P$ (sign opposite to LDA, orienting $P$ toward maximal within-pair variance relative to total). $L_\mathcal{F} = P^\top C_W P$ then has maximal spread, the spectral $H^1$ complement is more selectively concentrated, and $H^0$ picks up residual content within the highly variable basis.

\textbf{Coboundary, Laplacian, decomposition.} Orienting each edge as $e = (v_2, v_1)$, with $h^{(1)}$ at $v_1$, the coboundary $(\delta^0 s)_e = P^\top s_{v_1} - P^\top s_{v_2}$ measures pair disagreement, and the consistency energy $E_{\mathrm{cons}}(s) = \sum_e \|(\delta^0 s)_e\|^2$ vanishes on $H^0 = \ker(\delta^0)$. The empirical sheaf Laplacian in edge space is
\begin{equation}
L_\mathcal{F} = \frac{1}{N}\sum_{i=1}^N \Delta_i^{(k)} (\Delta_i^{(k)})^\top \in \mathbb{R}^{k\times k}, \quad \Delta_i^{(k)} = P^\top h_i^{(1)} - P^\top h_i^{(2)}.
\label{eq:empirical_laplacian}
\end{equation}
This $k \times k$ matrix equals $\tfrac{1}{N} Y L_G Y^\top$ where $Y = [P^\top h_v]_{v}$ and $L_G$ is the Laplacian of the matching graph. It is the second-moment matrix of the edge values of $\delta^0 h$, not the $|V|d \times |V|d$ operator $(\delta^0)^\top \delta^0$, and we denote it by $L_\mathcal{F}$. Given a readout $w \in \mathbb{R}^k$, $w^\top L_\mathcal{F} w$ is the Laplacian quadratic form of the observed data seen through $w^\top P^\top$ \cite{hansen2019learning,gebhart2023knowledge}, and minimizing it is a locality preserving projection on the paraphrase graph \cite{he2003locality}. Its kernel is the set of readouts under which the data is a global section, i.e.\ $v \mapsto w^\top P^\top h_v$ is constant along all edges, or equivalently $w^\top \Delta_i^{(k)} = 0$ for all $i$. We define $H^0$ to be the span of the lowest $m$ eigenvectors of $L_\mathcal{F}$ (a low-energy approximation to $\ker L_\mathcal{F}$) and $H^1$ directions as the span of the highest $m$. The pullback to the hidden space is $\tilde v_j = P v_j$, and the $H^0$ projector is given by $\Pi_{H^0} = P V_{H^0} V_{H^0}^\top P^\top$ ($P$ has orthonormal columns for the joint-PCA map, so $\Pi_{H^0}$ is an orthogonal projector; the CCA and contrastive maps of \S\ref{sec:steering_diagnostic} have columns with unit norm but are not orthogonalised). The eigenvalue problem is $k \times k$ ($k = 128$) regardless of the hidden dimension, so this step costs the same for 124M models as it does for 13B.

\textbf{Steering.} Given a style direction $\Delta = \bar h_{\text{target}} - \bar h_{\text{source}}$, we form (a) $\Delta_{H^1} = \Delta - \Pi_{H^0}\Delta$ ($H^0$-removed) or (b) the projection of $\Delta$ onto the top $m$ $H^1$ dimensions weighted by $s_j = \sum_\ell (\tilde v_\ell)_j^2 \lambda_\ell$. All steering vectors are norm-matched to $\alpha = 0.3 \|\bar h\|$. Baselines: random, full style (contrastive activation addition, CAA~\cite{rimsky2024steering}), mean-of-differences, variance top-$k$, Fisher LDA.

\textbf{The matching-graph limitation.} The construction so far gives local consistency: $\delta^0$ measures within-pair disagreement and $H^0$ contains its kernel. The graph itself, however, has no cycles, and $P^\top$ is surjective ($P$ has full column rank), so the cohomology vanishes above degree zero.

\begin{proposition}[Reduction to projected within-pair covariance]
\label{prop:laplacian_reduction}
With $C_W = \frac{1}{N}\sum_i \Delta_i \Delta_i^\top$ and $C_T = \frac{1}{N}\sum_i h_i h_i^\top$, on a matching graph with one $P$ shared by every node and identity transports on edges: (i) $L_\mathcal{F} = P^\top C_W P$; (ii) the Fisher baseline is the generalized eigenproblem $L_\mathcal{F} v = \lambda (P^\top C_T P) v$; (iii) $\lambda = \frac{1}{N}\sum_i (\Delta_i^\top P v)^2$ for a unit eigenvector $v$. Proof: Appendix~\ref{app:cohom_validation}.
\end{proposition}

\begin{remark}[Per-node restriction maps]
\label{rem:pernode}
With per-node maps $P_v$ and a Procrustes transport $Q_{uv}$ (the orthogonal map that best aligns stalk $u$ to stalk $v$), the coboundary is $(\delta^0 x)_e = P_v^\top x_v - Q_{uv} P_u^\top x_u$, and $L_\mathcal{F}$ is in general not $P^\top C_W P$, nor $P^\top M P$ for any $M$ built from hidden-state second moments alone, because the cross-terms $-P_v^\top h_v h_u^\top P_u Q_{uv}^\top$ carry different maps on the two sides.
\end{remark}

\begin{remark}[Terminology: $H^1$ directions]
\label{rem:h1}
Matching graphs are acyclic, so algebraic $H^1 = \ker(\delta^1)/\mathrm{im}(\delta^0) = 0$. We call the top eigenspace of $L_\mathcal{F}$ the \emph{$H^1$ directions} as a mnemonic for high disagreement; it is not a cohomology class. On a matching graph every 1-cochain, $\delta^0 h$ included, is a coboundary, so the name refers to where $\delta^0 h$ carries the most energy. Algebraic $H^1 \neq 0$ appears only on the cycle-bearing topologies of \S\ref{sec:hypergraph}.
\end{remark}

Proposition~\ref{prop:laplacian_reduction} needs a single $P$ shared across nodes with identity transports. It holds on any graph, and we apply it on the matching graph. Within that regime, the restriction-map ablation (Table~\ref{tab:restriction_ablation}) places the content in the kernel of the within-pair difference covariance whatever $P$ is, since random or identity maps recover most of the $H^0$ retrieval; empirically the Fisher baseline of part (ii) degrades fact preservation ($15.4\%$ vs.\ $29.8\%$). Outside it, multi-layer grids ($b_1 = N(L-1)$) carry a model-derived harmonic mass (\S\ref{sec:multilayer}), and per-node $P_v$ with Procrustes transports break the reduction (Remark~\ref{rem:pernode}, Appendix~\ref{app:holonomy_null}).

\section{Multi-Layer Matching Grids and Algebraic $H^1$}
\label{sec:hypergraph}

A matching graph consisting of one layer only gives the trivial answer for algebraic $H^1$, as there are no closed loops and $P^\top$ is onto. Filling in the cliques of mutually paraphrasing sentences is of no help, as these are all simply connected and hence $H^1$ vanishes (unlike $H^2$). The most direct source of closed loops is the number of layers.

\subsection{Multi-layer matching grids}
\label{sec:multilayer}

Take the vertex set to be $V = \{(s, l): s \in \mathrm{sentences}, l \in \mathrm{layers}\}$. There are edges between the two paraphrases of a pair within each layer, and edges between the same sentence in successive layers. This 1-complex is thus a rectangular mesh: if there are $N$ paraphrase pairs and $L$ layers it has $|V| = 2NL$, $|E| = NL + 2N(L-1)$ and no 2-faces. Direct computation gives $b_1 = N(L-1)$ independent algebraic 1-cycles. Each elementary cycle is a square (sentence $s$, paraphrase $s'$) at layers $l$ and $l{+}1$, and asks if applying the layer-$l \to l{+}1$ map commutes with passing to a paraphrase. If so, the sum of $\omega$ over the 4 sides of the square will be 0, that is, $\omega$ restricted to that cycle is exact and will have no harmonic part along the dual harmonic 1-cochain; otherwise there will be a non-zero projection of $\omega$ onto this harmonic 1-cochain.

\begin{table}[t]
\centering
\caption{Multi-layer matching grid, $b_1 = N(L{-}1)$ from depth alone. Hodge harmonic mass of model-derived edge cochains varies by nearly three orders of magnitude. Every row shares the same graph ($b_1 = 30$ at $L = 2$), so the variation comes from the model's layer transitions. $30$ MRPC pairs with joint-PCA per layer at shared $k = 8$. The last column is steering fragility: CounterFact fact preservation (\%) under random steering (\S\ref{sec:fragility}, Appendix~\ref{app:fragility}).}
\label{tab:multilayer}
\small
\begin{tabular}{@{}lccccc@{}}
\toprule
\textbf{Architecture} & $L=2$ & $L=4$ & $L=8$ & $L=16$ & \textbf{Pres.\ (\%)} \\
\midrule
Mistral-7B    & $0.0001$ & $0.0001$ & $0.0004$ & $0.0008$ & $39.4$ \\
Qwen-2.5-7B   & $0.0001$ & $0.018$  & $0.016$  & $0.096$  & $34.3$ \\
Llama-3-8B    & $0.001$  & $0.002$  & $0.003$  & $0.004$  & $36.0$ \\
Phi-2         & $0.033$  & $0.126$  & $0.109$  & $0.067$  & $27.5$ \\
GPT-2 (124M)  & $0.036$  & $0.107$  & $0.083$  & --       & $9.1$ \\
Llama-2-7B    & $0.079$  & $0.071$  & $0.047$  & $0.028$  & $4.2$ \\
\bottomrule
\end{tabular}
\end{table}

\begin{figure}[t]
\centering
\includegraphics[width=0.55\linewidth]{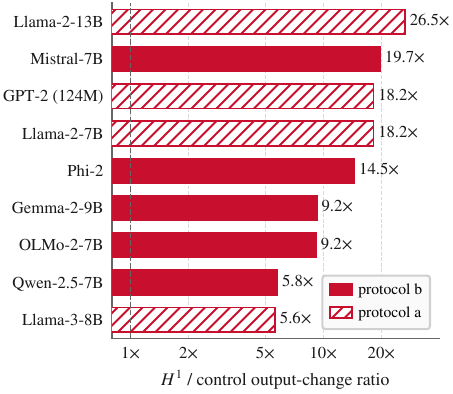}
\caption{Per-architecture ratio of the output change from ablating the $H^{1}$ coordinates to that from ablating control coordinates of matched per-coordinate variability (Table~\ref{tab:causal_nine_arch}). Hatched bars use protocol $a$ (one layer, $64$ coordinates, effect on logits) and solid bars protocol $b$ (three depths, $20$ coordinates, effect on the final hidden state). Every architecture exceeds the dashed unit baseline, by $5.6\times$ to $26.5\times$.}
\label{fig:causal_validation}
\end{figure}

\textbf{Definition of the model-derived edge cochain $\omega$.} There is a joint-PCA restriction map $P_l^\top$ for each layer $l$ ($P_l \in \mathbb{R}^{d \times k}$), applicable to every sentence. Denote by $h_s^l$ the mean-pooled hidden representation of sentence $s$ at layer $l$. Then an edge along the horizontal, joining the 2 paraphrases $a, b$ at layer $l$, has value $\omega(e) = P_l^\top (h_b^l - h_a^l)$. And an edge vertically, corresponding to sentence $s$ and going from layer $l$ to layer $l{+}1$, has value $\omega(e) = (P_{l+1}^\top h_s^{l+1} - \mu_{l+1}) - R_{l \to l+1}^\top (P_l^\top h_s^l - \mu_l)$, with $\mu_l$ denoting the mean over all projected activations at layer $l$ and $R_{l \to l+1}$ the orthogonal Procrustes solution mapping the centred projections of the $2N$ sentences at one layer to the other. If $R = I$, then (up to centring) $\omega$ would be the coboundary of a node potential, with zero harmonic mass. So the mass varies according to the model through the fitted transport. Stacking these gives a single $\omega \in \mathbb{R}^{|E| \cdot k}$, and the Hodge harmonic mass is $\|\Pi_{\ker L_1} \omega\|^2 / \|\omega\|^2$, with $L_1 = \delta^0 (\delta^0)^\top + (\delta^1)^\top \delta^1$ the Hodge 1-Laplacian, equal to the usual edge Laplacian $\delta^0 (\delta^0)^\top$ since there are no 2-cells. With all edge stalks identified with one $\mathbb{R}^k$, this amounts to taking the Hodge decomposition of the constant sheaf $\mathbb{R}^k$, applied to $\omega$ separately in each coordinate \cite{hansen2019toward,schaub2021signal}. Noting that $\omega$ is the coboundary of the node potentials $P_l^\top h_s^l - \mu_l$ transported by $R_{l \to l+1}$, only the action of $I - R_{l \to l+1}^\top$ on differences within each pair of sentences can contribute harmonically. The mass is evaluated in the coordinates given by the joint-PCA maps $P_l$. The decomposition $C^1 = \mathrm{im}\,\delta^0 \oplus \ker L_1 \oplus \mathrm{im}\,(\delta^1)^\top$ is orthogonal \cite{lim2020hodge}. The effect of varying $k$ and the choice of restriction maps is discussed in Appendix~\ref{app:simplicial_scaleup}; the division into robust and fragile models continues to hold with $k \in \{8, 64, 128\}$.
\smallskip

Since $b_1$ grows from 0 linearly with $L$, depth alone determines the number of cycles, and the harmonic mass given by the model ranges over $[10^{-4}, 0.08]$ at $L = 2$ (Table~\ref{tab:multilayer}). A randomly drawn Gaussian cochain would have on average a harmonic fraction of exactly $b_1/|E| = 0.25$ at $L = 2$, providing a useful benchmark. Almost none (about $10^{-4}$) of the mass of the model's $\omega$ lies on the space of harmonics for both Mistral-7B and Qwen-2.5-7B, but a substantial amount does for Llama-2-7B. The split between the three low-mass and three high-mass models is unchanged if $k$ is increased from $8$ to $128$ and all squares are triangulated (Appendix~\ref{app:simplicial_scaleup}). Section~\ref{sec:fragility} asks whether the mass is indicative of steering fragility.

\textbf{Why the 2-simplices stay unfilled.} There are no 2-cells, hence no cohomology above degree 1. Taking coefficients in $\mathbb{R}$, $\dim H^1 = b_1 = N(L-1)$ ($k\,b_1$ with $\mathbb{R}^k$ stalks). If we fill in each square with 2 simplices, meeting along a common diagonal $(a, l) \mathchar`- (b, l+1)$, then each square cycle is now the boundary of $T_1 - T_2$, oriented according to the ordering of the vertices. Direct computation on this 2-complex gives $b_1 = b_2 = 0$, for all values of $L$ and $N$ tried. Hence there is no harmonic mass. What is left is the coexact fraction of $\omega$, which measures the circulation of $\omega$ around the filled triangles, and still distinguishes robust models from fragile ones (Appendix~\ref{app:simplicial_scaleup}).

\subsection{Relation hypergraphs}
\label{sec:relation_hypergraph}

An alternative way to obtain cycles is based on relations rather than depth. Let $\mathcal{D}$ contain $n_f$ facts, each with $n_p$ paraphrases, grouped under $n_r$ relations. Put a ring around the $n_p$ paraphrases of every fact, and one through the first paraphrase of all facts belonging to each relation, and fill in no 2-simplices. For $n_f = 60$, $n_p = 5$, $n_r = 6$ this results in $300$ nodes and $300 + 60 = 360$ edges with no faces, with $\dim H^0 = 6$ and $\dim H^1 = 66$ (Euler $\chi = -60 = 6 - 66$). The $66$ classes are made up of $60$ rings around individual facts and $6$ cycles around relations. The matching graph on the same nodes has $\dim H^1 = 0$, and if all 2-simplices corresponding to individual facts are added it is again $\dim H^1 = 0$. Thus the non-trivial topology is due only to the relation hypergraph, for the same activations.

\textbf{Heterogeneous restriction maps and holonomy.} Computing $P_v$ for each node separately, with Procrustes transports $Q_{uv} \neq I$, breaks Proposition~\ref{prop:laplacian_reduction} (Remark~\ref{rem:pernode}). All cycle holonomy obstructions on the relation hypergraph are $2.5$--$3.1$, below the expected RMS under Haar $\sqrt{2k} = 4$, and no model deviates significantly from the levels expected with randomly shuffled paraphrases (one-sided $p = 0.10$--$0.33$, Appendix~\ref{app:holonomy_null}). We were also unable to use $H^1$ in node space to improve retrieval or steering (Appendix~\ref{app:h1_node_space}). So both constructions with cycles, namely grids from depth and hypergraphs from relations, move the framework off the matching-graph $H^1 = 0$ floor.

\section{Experiments}
\label{sec:experiments}

\subsection{Setup and Datasets}

We test 9 transformer models of varying architectures (GPT-2~\cite{radford2019language}, Llama~\cite{touvron2023llama2,grattafiori2024llama3}, Mistral~\cite{jiang2023mistral}, Qwen~\cite{yang2024qwen2}, phi~\cite{li2023phi15}, Gemma~\cite{gemmateam2024gemma2}, OLMo~\cite{olmo2025olmo2}) and sizes (124M to 13B parameters). Functional validation ablates coordinates over $1000$ sentences (4 models; $666$ MRPC and $334$ PAWS) or $100$ MRPC sentences (5 models). Restriction maps (PCA, CCA, contrastive Fisher) are learned from 200 pairs of the Microsoft Research Paraphrase Corpus (MRPC)~\cite{dolan2005mrpc}. We steer using CounterFact~\cite{meng2022locating} ($n = 1000$ per model) and retrieve facts using ROME CounterFact, with 500 facts split 70/30 for held-out evaluation (\S\ref{sec:leace_comparison}) and a 60-fact, 5-paraphrase subset for the topology constructions. We use mean-pooled hidden states from layer $\ell \approx 2L/3$, set the edge dimension to $k = 128$ to capture $\geq 98\%$ of the paired variance, take the lowest and highest 20 eigenvectors as the $H^0$ and $H^1$ directions, and norm-match all steering vectors to $\alpha = 0.3 \|\bar{h}\|$. We use bootstrap CIs with 1000 resamples, McNemar's test~\cite{mcnemar1947note} for paired comparisons, and Cohen's $d$~\cite{cohen1988statistical} for effect sizes.

\subsection{Content from the Kernel: \texorpdfstring{$H^0$}{H0} Beats LEACE, and Joint Paraphrase Constraints Improve It}
\label{sec:content}\label{sec:leace_comparison}\label{sec:fact_probe}

Paraphrase-invariance alone does not guarantee fact discriminability: an invariant subspace could conflate Paris--France with Berlin--Germany. Hard same-relation retrieval tests this, against baselines that are low-dimensional or designed for invariance. We compare sheaf $H^0$ (20D) to LEACE's preserved subspace on ROME CounterFact, with a 70/30 train/test split of 500 facts ($n_{\text{test}}{=}150$). LEACE erases the binary paraphrase/non-paraphrase concept with an oblique rank-$(d-1)$ projection $P^*$, and we call $\mathrm{im}\,P^*$ the LEACE-preserved subspace, at full hidden dim, while sheaf $H^0$ is $20$-dimensional and built without binary labels. Easy retrieval is top-1 over all held-out centroids, and hard retrieval is limited to same-relation distractors. Other subspace and metric-learning baselines under the identical protocol are in Appendix~\ref{app:baselines}: $H^0$ outperforms every label-free alternative and performs worse than LDA and NCA with fact labels on 5/8 models.

\begin{table}[t]
\centering
\caption{Held-out CounterFact retrieval (top-1 accuracy, hard same-relation distractors, $k=20$ dims, 70/30 train/test split, $n_{\text{test}} = 150$ facts). Sheaf $H^0$ basis and LEACE eraser are fit on training facts only, and evaluation is on held-out facts not seen during fitting. Across eight architectures sheaf $H^0$ at 20 dimensions beats LEACE-preserved at full hidden dimension, with gaps ranging from $+1.5$pp (OLMo-2) to $+42.2$pp (GPT-2).}
\label{tab:leace_held_out}
\small
\begin{tabular}{@{}lccccccc@{}}
\toprule
\textbf{Model} & \textbf{Full} & $H^0$ \textbf{(20D)} & \textbf{LEACE-full} & \textbf{LEACE-20D} & \textbf{PCA-20} & \textbf{Rand-20} & \textbf{$\Delta$} \\
\midrule
GPT-2          & 43.1 & \textbf{85.3} & 43.1 & 28.9 & 24.6 & 39.1 & +42.2 \\
Mistral-7B     & 77.7 & \textbf{87.1} & 77.7 & 36.8 & 34.5 & 59.2 & +9.4 \\
OLMo-2-7B      & 81.3 & \textbf{82.8} & 81.3 & 40.9 & 35.7 & 62.3 & +1.5 \\
Phi-2          & 69.1 & \textbf{87.1} & 69.1 & 38.8 & 37.2 & 55.1 & +18.0 \\
Qwen-2.5-7B    & 49.4 & \textbf{64.1} & 49.4 & 36.8 & 37.5 & 43.7 & +14.7 \\
Gemma-2-9B     & 54.4 & \textbf{77.9} & 54.4 & 32.3 & 27.5 & 43.5 & +23.5 \\
Llama-2-7B     & 68.2 & \textbf{80.4} & 68.2 & 25.7 & 22.3 & 50.7 & +12.2 \\
Llama-3-8B     & 72.7 & \textbf{88.0} & 72.7 & 28.2 & 24.4 & 55.9 & +15.3 \\
\bottomrule
\end{tabular}
\end{table}

Hard same-relation retrieval (Table~\ref{tab:leace_held_out}) shows sheaf $H^0$ with 20 dimensions beats LEACE-preserved at full hidden dim on all 8 models (mean $+17.1$pp). The $+42.2$pp GPT-2 gap matches the smallest model, since smaller residual streams concentrate paraphrase content into lower-dimensional shared subspaces. Even on the 7--9B models, $H^0$ with $20$ dimensions outperforms LEACE with $3584$--$4096$ dimensions by $1.5$--$23.5$pp.

\textbf{Dimension-controlled LEACE.} LEACE for a binary concept removes a single direction (its null space has rank $1$, so the eraser $P^*$ has rank $d-1$) \cite{belrose2023leace}, so the dim-controlled $20$D baseline (LEACE-20D, Table~\ref{tab:leace_held_out}) is the top-$20$ PCs of the LEACE-preserved activations. Sheaf $H^0$ beats LEACE-20D by $27$--$60$\,pp on every architecture (mean $+48.0$\,pp). As expected from the algebra, LEACE-20D gives results within $6$\,pp of those for PCA-20, that is, a rank-1 erasure has little effect on the ranking of the top PCs. So at matched dimensionality the gap can be ascribed to the decomposition.

\textbf{The result is down to the kernel selection, with the projection $P$ interchangeable.} Table~\ref{tab:restriction_ablation} ablates $H^0$ on the same activations under three $P$ regimes: PCA-$P$ (current), random orthonormal $P$ (5 seeds), and identity $P$ (bottom-20 eigenvectors of $C_W$ in full $d$-dim native space). Mean across eight architectures: PCA-$P$ $81.6\%$, random-$P$ $74.2\%$, identity-$P$ $77.5\%$ vs.\ PCA-20-without-kernel $30.5\%$ and Random-20 $51.2\%$. For identity $P$, this is simply the bottom eigenspace of $C_W$ itself. Thus, for any $P$, the content is carried by the kernel of the covariance over differences within pairs, and using PCA to solve the eigenproblem in $\mathbb{R}^{128}$ rather than $\mathbb{R}^d$ incurs a penalty of at most $7.4$\,pp. Non-learned restriction maps likewise match learned ones on heterophilic-graph benchmarks~\cite{hernandez2024joint}; closed-form learned restriction maps~\cite{dinino2025learning} are complementary.

\textbf{Joint paraphrase constraints improve $H^0$.}\label{sec:hypergraph_cohomology} The result above uses a pair-only graph, in which each fact contributes a chain of paraphrase pairs. Replacing it with a ring over all of a fact's paraphrases ($b_1 = N$ instead of $0$), with activations and $P$ unchanged, asks for invariance across every paraphrase of a fact jointly. Across eight architectures ring-$H^0$ beats pair-$H^0$ by $+3.2$ to $+9.5$\,pp (mean $+6.5$\,pp, Table~\ref{tab:cycle_h0}); a clique matches the ring within $0.1$\,pp. An edge-count-matched control attributes the gain to the joint constraint: an acyclic path over six paraphrases, with the ring's five edges and $b_1 = 0$, recovers it ($+7.7$\,pp over pairs), and the ring exceeds that path by $+0.5$\,pp, significant on none of the eight architectures (Appendix~\ref{app:cycle_h0}). The cycle-closing difference is the sum of the path differences, so it adds no constraint on exact global sections: $H^0$ is set by which paraphrases are connected, and cycles enter through the harmonic subspace of \S\ref{sec:multilayer}.

\subsection{Influence from the Spectral Complement: \texorpdfstring{$H^1$}{H1} Carries Concentrated Causal Influence}
\label{sec:influence}\label{sec:causal_validation}

We ablate the $H^1$ directions using a coordinate-axis proxy and compare to controls. At layer $\ell$ we rank the $d$ residual-stream coordinates by mean absolute difference within pairs $\mathbb{E}|h^{(1)}_{i,j} - h^{(2)}_{i,j}|$ on MRPC. This is the diagonal analogue of $C_W$ and thus a proxy for the top eigenspace of $L_\mathcal{F}$ with the identity as the restriction map (Table~\ref{tab:restriction_ablation} shows this map recovers most of the $H^0$ result). The top $m$ coordinates are $H^1$ and the bottom $m$ are $H^0$. The controls are the $m$ coordinates in neither set with individual variability closest to the average variability of the $H^1$ coordinates. We zero the chosen coordinates in the layer-$\ell$ output, continue the forward pass, and measure the $L_2$ change in the model's output: the logit change at the last position for the $1000$-sentence runs ($m = 64$, one layer at $L/2$) and the change in the mean-pooled final hidden representation for the $100$-sentence runs ($m = 20$, layers $L/4$, $L/2$, $3L/4$). The ratio is the mean change under $H^1$ ablation over the mean change under control ablation, with a bootstrap CI. The controls are matched coordinate to coordinate and have only $0.13$ to $0.53$ of the total variability of the $H^1$ set. So the ratio compares coordinates of matched variability, and part of it reflects the greater total variability that the $H^1$ set removes~\cite{makelov2024subspace,wu2024reply}. We use ``causal'' in the activation patching sense \cite{vig2020causal,geiger2024finding,chan2022scrubbing}. Projecting out the sheaf $H^1$ eigenbasis itself causes $17.8$--$398\times$ more change in the next-token distribution than projecting out a random subspace of the same size across all 9 models (median over layers, Appendix~\ref{app:sheaf_basis}). It also affects the generated text: fact recall collapses under $H^1$ ablation and stays at baseline under matched random ablations. The $H^1$ subspace is the least dormant subspace we have measured, which argues against the dormant pathway explanation of the subspace patching illusion~\cite{makelov2024subspace} (Appendix~\ref{app:illusion}).

\begin{table}[t]
\centering
\caption{Functional validation: output change from ablating the $H^1$ coordinates over that from ablating control coordinates of matched per-coordinate variability. $^{a}$\,$1000$ sentences ($666$ MRPC, $334$ PAWS), $64$ coordinates, one layer ($L/2$), effect on final-position logits. $^{b}$\,$100$ sentences, $20$ coordinates, three layers ($L/4$, $L/2$, $3L/4$), effect on the mean-pooled final hidden state, best of the three depths shown. CIs are bootstrap $95\%$. Cohen's $d$ uses the pooled standard deviation of the two effect distributions.}
\label{tab:causal_nine_arch}
\small
\begin{tabular}{@{}lcccc@{}}
\toprule
\textbf{Model} & \textbf{Ratio} & \textbf{Layer} & \textbf{95\% CI} & \textbf{Cohen's $d$} \\
\midrule
GPT-2 (124M)$^{a}$ & 18.2$\times$ & L6 & [17.2, 19.3] & 2.34 \\
Phi-2$^{b}$ & 14.48$\times$ & L8 & [14.0, 15.0] & 13.98 \\
Mistral-7B$^{b}$ & 19.71$\times$ & L8 & [19.1, 20.3] & 18.45 \\
Qwen-2.5-7B$^{b}$ & 5.76$\times$ & L7 & [5.38, 6.12] & 6.63 \\
OLMo-2-7B$^{b}$ & 9.17$\times$ & L16 & [8.85, 9.49] & 12.93 \\
Gemma-2-9B$^{b}$ & 9.24$\times$ & L21 & [8.94, 9.54] & 22.02 \\
Llama-2-7B$^{a}$ & 18.2$\times$ & L16 & [17.3, 19.1] & 4.44 \\
Llama-3-8B$^{a}$ & 5.62$\times$ & L16 & [5.52, 5.71] & 9.08 \\
Llama-2-13B$^{a}$ & \textbf{26.5$\times$} & L20 & [26.1, 26.8] & \textbf{14.3} \\
\bottomrule
\end{tabular}
\end{table}

The $H^1$ coordinates have 5.6--26.5$\times$ more influence than the control coordinates on all 9 models, with all 95\% CIs excluding 1 (Table~\ref{tab:causal_nine_arch}, Figure~\ref{fig:causal_validation}). GPT-2 (124M) and Mistral-7B have near-identical ratios (18.2$\times$ and 19.71$\times$) despite being 56$\times$ apart in size. At 3 depths on all 9 models (protocol $b$), the ratio peaks at $L/4$ for 7 models and at $L/2$ for OLMo-2-7B and Gemma-2-9B (Appendix~\ref{app:layer_dynamics}).

\subsection{Fragility from Harmonic Mass: Sheaf Steering as an Architecture Diagnostic}
\label{sec:fragility}\label{sec:steering_diagnostic}

Harmonic mass is computed from MRPC layer-transition differences and steering fragility from CounterFact fact preservation under random perturbation, so the two come from disjoint measurements of the same model. Fragility is the fraction of CounterFact facts preserved under a random direction norm-matched to $0.3$ times the hidden-state norm at layer $\lfloor 5L/8 \rfloor$ (Appendix~\ref{app:fragility}). The three lowest-mass architectures are the three most robust and the three highest are the three most fragile: Spearman $\rho = 0.90$, exact one-sided permutation $p = 0.014$ over the $720$ rankings. The one rank inversion, Qwen-2.5-7B against Llama-3-8B, is within sampling error ($34.3\%$ vs.\ $36.0\%$). Table~\ref{tab:multilayer} lists both quantities per model.

The rest of this subsection compares steering methods within a model. CounterFact evaluates whether steering preserves facts while changing style. We compute style direction as the difference in mean activations between target and original completions at layer $\ell = \lfloor 5L/8 \rfloor$, then add norm-matched perturbations. Style change is measured as the fraction of outputs differing from baseline, and fact preservation is the fraction containing the target entity (greedy decoding, 50 tokens). At $\alpha = 0.3$, style saturates at 100\% for all methods on all models, so fact preservation is the key differentiator.

\begin{table}[t]
\centering
\caption{Steering on CounterFact ($\alpha=0.3$, $n=1000$ per model). Fact preservation (\%) with 95\% bootstrap CIs. McNemar $p$ is paired against the best method in each block (bold). Style saturates at 100\% for all methods on all models.}
\label{tab:steering_four_arch}
\small
\begin{tabular}{@{}llccc@{}}
\toprule
\textbf{Model} & \textbf{Method} & \textbf{Fact\%} & \textbf{95\% CI} & \textbf{McNemar $p$} \\
\midrule
\multirow{6}{*}{Llama-2-7B}
& Contrastive sheaf ($H^1$ dims) & \textbf{31.0} & [28.2, 33.9] & -- \\
& PIXEL~\cite{yu2025pixel} & 21.1 & [18.6, 23.9] & $< 10^{-16}$ \\
& Full style (CAA) & 13.7 & [11.7, 15.9] & $< 10^{-31}$ \\
& $H^0$-removed (sheaf) & 13.7 & [11.6, 15.9] & $< 10^{-31}$ \\
& $H^1$ dims (sheaf PCA) & 12.1 & [10.0, 14.1] & $< 10^{-38}$ \\
& Random & 4.2 & [3.0, 5.4] & $< 10^{-70}$ \\
\midrule
\multirow{5}{*}{Llama-3-8B}
& PIXEL~\cite{yu2025pixel} & \textbf{38.8} & [35.2, 42.4] & -- \\
& Random & 36.0 & [33.0, 38.8] & 0.10 \\
& Full style (CAA) & 27.2 & [24.5, 29.9] & $< 10^{-15}$ \\
& $H^1$ dims (sheaf PCA) & 26.1 & [23.4, 28.7] & $< 10^{-2}$ \\
& $H^0$-removed (sheaf) & 26.1 & [23.3, 28.8] & $< 10^{-2}$ \\
\midrule
\multirow{6}{*}{Mistral-7B}
& CCA sheaf ($H^1$ dims) & \textbf{41.6} & [38.6, 44.4] & -- \\
& Random & 39.4 & [36.3, 42.4] & 0.19 \\
& PIXEL~\cite{yu2025pixel} & 34.8 & [31.1, 38.3] & $< 10^{-8}$ \\
& Variance (top-$k$) & 33.0 & [30.0, 35.8] & $< 10^{-7}$ \\
& Full style (CAA) & 31.7 & [28.8, 34.6] & $< 10^{-9}$ \\
& $H^1$ dims (sheaf PCA) & 29.8 & [26.9, 32.6] & $< 10^{-12}$ \\
\midrule
\multirow{4}{*}{Llama-2-13B}
& $H^1$ dims (sheaf PCA) & \textbf{32.7} & [29.8, 35.6] & -- \\
& Random & 32.1 & [29.3, 35.0] & 0.74 \\
& Variance (top-$k$) & 31.3 & [28.5, 34.1] & 0.82 \\
& Full style (CAA) & 29.6 & [26.8, 32.3] & 0.07 \\
\bottomrule
\end{tabular}
\end{table}

Table~\ref{tab:steering_four_arch} shows an architecture-dependent pattern. Llama-2-7B is fragile, and random steering collapses fact generation (4.2\%), while directed methods retain facts at 12--31\%, led by the architecture-matched contrastive sheaf at 31.0\% (McNemar $p < 10^{-70}$ vs.\ random). Mistral-7B is robust, and the architecture-matched CCA sheaf has the highest value (41.6\% vs.\ 39.4\% for random, McNemar $p = 0.19$, so they are statistically tied). Llama-3-8B is also robust, with random and PIXEL having the highest values (36.0--38.8\%). Random can compete at higher $\alpha$ on robust models, which is consistent with the concentrated influence of the $H^1$ coordinates (\S\ref{sec:influence}) disrupting more at large magnitudes. Llama-2-13B falls in between, all methods performing similarly (29.6--32.7\%, McNemar $p > 0.07$). Consistent with this, the sheaf directions outperform random at lower magnitudes (see below).

\textbf{Operating regime.} We present $\alpha = 0.30$ as a stress test. Over a sweep of $\alpha \in [0.05, 0.50]$ (Appendix~\ref{app:scale_sweep}), $H^0$-removed sheaf steering with the architecture-matched map outperforms random on both swept models (Mistral-7B and Llama-3-8B) at $\alpha \in [0.10, 0.15]$ (averaging $+1.0$--$2.0$\,pp higher, with $H^0$-removed reaching $43\%$ vs.\ $41\%$ for random on Mistral-7B at $\alpha=0.10$), and random becomes competitive from $\alpha = 0.30$. Use $\alpha \in [0.10, 0.15]$ on robust models and $\alpha \approx 0.30$ on fragile ones.

\textbf{Head-to-head with PIXEL.} PIXEL~\cite{yu2025pixel}, which extracts steering directions from two views of the prompt and calibrates the magnitude per position, reproduces the architecture-dependent pattern (Table~\ref{tab:steering_four_arch}): $21.1\%$ on Llama-2-7B (vs.\ $4.2\%$ random) but $34.8\%$ on Mistral-7B where random reaches $39.4\%$. PIXEL beats CAA on Llama-2-7B ($p < 10^{-7}$), and contrastive sheaf leads both at $31.0\%$. The fragility pattern holds for both direction recipes, and our simplifications of PIXEL appear in Appendix~\ref{app:pixel_comparison}.

\section{Related Work}
\label{sec:related}

This paper extends a non-archival workshop version~\cite{grant2026gluingworkshop} that contained only the single-layer construction with three pilot experiments; the multi-layer grids, relation hypergraphs ($b_1 = 66$), harmonic-mass fragility diagnostic, held-out LEACE protocol, restriction-map ablation, PIXEL comparison, baseline suite, and self-supervised learning (SSL) connection are new.

\textbf{Sheaves in machine learning.} Hansen and Ghrist~\cite{hansen2019toward} developed the spectral theory of the sheaf Laplacian, which specializes to the connection Laplacian of Singer and Wu~\cite{singer2012vector} on $O(d)$-bundles, and Hansen and Gebhart~\cite{hansen2020sheaf} introduced sheaf neural networks, since applied to heterophilic graphs~\cite{bodnar2022neural,barbero2022sheaf,hernandez2024joint}, predictive coding~\cite{seely2025sheaf}, and restriction-map learning~\cite{dinino2025learning}. Duta et al.~\cite{duta2023sheaf} build sheaf hypergraph networks, and Mule et al.~\cite{mule2026directional} unify directed/undirected hypergraph Laplacians. In the closest concurrent work on transformers, Krasnovsky~\cite{krasnovsky2025eics} proposes a sheaf-inconsistency score for circuit coherence and Javidnia~\cite{javidnia2026gauge} builds context-chart atlases with Fisher metrics and holonomy. Our stalks are different (nodes of a paraphrase graph carrying hidden states under PCA/CCA/contrastive restrictions), as is our target, a post-hoc $H^0/H^1$ decomposition for retrieval and steering.

\textbf{Subspace interpretability.} Sparse autoencoders~\cite{bricken2023monosemanticity,cunningham2024sparse,templeton2024scaling,gao2025scaling}, probes~\cite{belinkov2022probing}, causal abstraction~\cite{geiger2021causal}, CKA/SVCCA~\cite{kornblith2019similarity,raghu2017svcca}, and distributed alignment search (DAS) and LEACE~\cite{geiger2024finding,belrose2023leace} target individual features or whole-space comparison. The $\log\det\mathrm{Cov}$ regulariser of \S\ref{sec:jepa} shares its geometry with the decorrelation terms of VICReg~\cite{bardes2022vicreg} and Barlow Twins~\cite{zbontar2021barlow}. Invariance objectives (contrastive: SimCLR~\cite{chen2020simple}, InfoNCE~\cite{oord2018representation}; non-contrastive: BYOL~\cite{grill2020byol}, Barlow Twins~\cite{zbontar2021barlow}; DCCA~\cite{andrew2013deep}) are training-time analogues that achieve paired-view agreement during training, while we read invariance off a frozen model via $\ker(L_\mathcal{F})$.

\textbf{Steering and linear representations.} Activation addition~\cite{turner2023activation}, representation engineering~\cite{zou2023representation} and their variants~\cite{yu2025pixel,vu2025angular,jiang2025msrs} add directions and are compared in a unified evaluation by Im and Li~\cite{im2025unified}. Da Silva et al.~\cite{dasilva2025steering} show across 36 models that steering reliability depends on the model, and Tan et al.~\cite{tan2025analysing} show that it's unreliable across inputs, which \S\ref{sec:fragility} connects to harmonic mass. The linear representation hypothesis~\cite{park2024linear,gurnee2024language,hernandez2024linearity} is qualified by Engels et al.~\cite{engels2025not} and Makelov et al.~\cite{makelov2024subspace}. Our per-coordinate variability-matched controls only partially address this concern (\S\ref{sec:influence}). We compare to PIXEL (\S\ref{sec:steering_diagnostic}, Appendix~\ref{app:pixel_comparison}). Angular Steering~\cite{vu2025angular} and MSRS~\cite{im2025unified} are application-side and orthogonal.

\section{Discussion}
\label{sec:discussion}

Proposition~\ref{prop:laplacian_reduction} reduces $L_\mathcal{F}$ to projected covariance for a single shared $P$ with identity transports, here on the matching graph. Within that regime the restriction-map ablation locates content in the kernel of the within-pair difference covariance (Table~\ref{tab:restriction_ablation}). Beyond it, multi-layer grids carry a model-derived harmonic mass whose ranking tracks steering fragility (\S\ref{sec:fragility}). The mass needs only 30 MRPC pairs and one forward pass per sentence, so it can screen a model before a steering sweep. Graphs that join activations of different dimension (text and image encoders, or models of different width) need per-node maps $P_v$ of different widths, which we leave to future work.

\textbf{Self-supervised connection.}\label{sec:jepa} A sheaf-regularized loss $\mathcal{L} = \mathrm{tr}(L_\mathcal{F}) - \beta \log\det\mathrm{Cov}(f)$ combines VICReg's invariance term ($\mathrm{tr}\,L_\mathcal{F}$ is the within-pair MSE up to scale) with a log-det spread term. Like VICReg's variance and covariance terms \cite{bardes2022vicreg} and Barlow Twins' redundancy reduction \cite{zbontar2021barlow}, this penalizes dimensional collapse, though not in VICReg's hinge-plus-off-diagonal form. Such regularizers are used alongside non-contrastive losses such as BYOL and SimSiam \cite{grill2020byol,chen2021simsiam}. Across 7 architectures it holds encoder rank at $57$--$60$/$64$, where MSE alone collapses to $14$--$42$/$64$ (Appendix~\ref{app:collapse}).

\textbf{Limitations.} $P$ is architecture-dependent and partially domain-dependent (Appendix~\ref{app:crossval}). The operating regime is $\alpha \in [0.10, 0.15]$. Node-space $H^1$ does not transfer, and cycle holonomy is not significantly different from a shuffled null (Appendices~\ref{app:h1_node_space},~\ref{app:holonomy_null}). Our fragility diagnostic is a rank correlation across 6 architectures ($\rho = 0.90$), and we would need more architectures to use it as a predictor. All the construction needs are contexts asserted to be equivalent to each other, and a cross-lingual probe shows it transfers to translation pairs (Appendix~\ref{app:crosslingual}). Logical equivalences and dialogue continuations fit the same template, but we have not tried them.

\section{Conclusion}
\label{sec:conclusion}

If one thinks of consistency across contexts as a sheaf condition, then every pair of paraphrases gives a local patch where the model agrees. Applying the kernel of the sheaf Laplacian yields a subspace of consistent content, while its spectral complement contains all variation dependent on the particular wording, for a fixed pre-trained network. On held-out retrieval, the kernel does better than the LEACE preserved subspace; the spectral complement captures much of the causal effect; and closed walks between different depths under this Laplacian define a harmonic mass, the relative values of which are predictive of susceptibility to steering, across six models ($\rho = 0.90$).

\section*{Acknowledgments}
This material is based upon work supported by the National Science Foundation Graduate Research Fellowship under Grant No.~2437833.

\bibliography{main}
\bibliographystyle{plainnat}

\newpage
\appendix

\section*{Appendix Contents}
\addcontentsline{toc}{section}{Appendix Contents}

{\small
\noindent
\begin{tabularx}{\linewidth}{@{}l X r@{}}
\multicolumn{3}{@{}l}{\textbf{\ref{app:theory}\quad Theoretical Foundations} \dotfill \pageref{app:theory}} \\[2pt]
\quad \ref{app:algebraic_h1}        & Algebraic $H^1 \neq 0$ on Richer Topologies   & \pageref{app:algebraic_h1} \\
\quad \ref{app:simplicial_scaleup}  & Simplicial Complex Variants and Scale-Up      & \pageref{app:simplicial_scaleup} \\
\quad \ref{app:fragility}           & Harmonic Mass vs Steering Fragility           & \pageref{app:fragility} \\
\quad \ref{app:geometric_picture}   & Geometric Picture: Where $H^1$ Lives          & \pageref{app:geometric_picture} \\
\quad \ref{app:cohom_validation}    & Proof of the Reduction and Spectral Validation & \pageref{app:cohom_validation} \\[4pt]
\multicolumn{3}{@{}l}{\textbf{\ref{app:experiments}\quad Additional Experiments} \dotfill \pageref{app:experiments}} \\[2pt]
\quad \ref{app:crossval}            & Cross-Dataset Generalization                  & \pageref{app:crossval} \\
\quad \ref{app:scale_sweep}         & Steering Magnitude Sweep                      & \pageref{app:scale_sweep} \\
\quad \ref{app:cycle_h0}            & Cycle-Aware $H^0$ (Multi-Paraphrase Rings)    & \pageref{app:cycle_h0} \\
\quad \ref{app:restriction_maps}    & Restriction Map Sensitivity                   & \pageref{app:restriction_maps} \\
\quad \ref{app:baselines}           & Retrieval Baseline Comparison                 & \pageref{app:baselines} \\
\quad \ref{app:sheaf_basis}         & Sheaf Eigenbasis Ablation                     & \pageref{app:sheaf_basis} \\
\quad \ref{app:illusion}            & Dormancy and the Subspace-Patching Illusion   & \pageref{app:illusion} \\
\quad \ref{app:crosslingual}        & Cross-Lingual Invariance Probe                & \pageref{app:crosslingual} \\
\quad \ref{app:layer_dynamics}      & Per-Architecture Layer Profiles               & \pageref{app:layer_dynamics} \\
\quad \ref{app:significance}        & McNemar Tests with BH Correction              & \pageref{app:significance} \\
\quad \ref{app:pixel_comparison}    & PIXEL Head-to-Head on CounterFact             & \pageref{app:pixel_comparison} \\
\quad \ref{app:holonomy_null}       & Holonomy Null Calibration                     & \pageref{app:holonomy_null} \\
\quad \ref{app:h1_node_space}       & Node-Space $H^1$ Operationalizations          & \pageref{app:h1_node_space} \\
\quad \ref{app:collapse}            & SSL Collapse Prevention (Per-Architecture)    & \pageref{app:collapse} \\
\quad \ref{app:mlst}                & MLST on STS-B (Full Table)                    & \pageref{app:mlst} \\[4pt]
\multicolumn{3}{@{}l}{\textbf{\ref{app:reproducibility}\quad Reproducibility} \dotfill \pageref{app:reproducibility}} \\[2pt]
\quad \ref{app:ablation}            & Hyperparameter Sensitivity                    & \pageref{app:ablation} \\
\quad \ref{app:details}             & Experimental Details                          & \pageref{app:details} \\
\quad \ref{app:practitioner}        & Practitioner Guidance                         & \pageref{app:practitioner} \\
\quad \ref{app:checklist}           & NeurIPS Paper Checklist                       & \pageref{app:checklist} \\
\end{tabularx}
}

\section{Theoretical Foundations}
\label{app:theory}

\subsection{Algebraic $H^1 \neq 0$ on Richer Topologies}
\label{app:algebraic_h1}

For the matching graph of the main text, $H^1 = 0$ algebraically, since there are no cycles and $P^\top$ is onto. We wish to construct examples with $H^1 \neq 0$, so consider the following 3 complexes, all built using the same 300 activation vectors (60 facts, 5 paraphrases each):

\begin{enumerate}
\item Matching graph (paper baseline): 300 nodes, 120 edges (one per paraphrase pair), 0 faces. Algebraic $H^1 = 0$.
\item Fact clique (filled): 300 nodes, 600 edges (all pairs), 600 filled triangles. All cliques are simply connected and so have $H^1 = 0$, although $H^2 = 240$.
\item Relation hypergraph (unfilled cycles): 300 nodes, 360 edges (a five-node ring around each fact, 300 edges, plus a ring for each relation through the first paraphrase of each of its ten facts, 60 edges), 0 filled faces. There are then loops which cannot be contracted, giving $\dim H^1 = 66$.
\end{enumerate}

The rings corresponding to relations link together the facts sharing a relation, for example, all \texttt{country-capital} facts. There are thus cycles in this complex which are not filled by any 2-simplex. The cohomology dimensions calculated this way are with trivial $\mathbb{R}$ coefficients on the underlying simplicial complex, and so need not agree with the sheaf cohomology of the induced learned-$Q$ sheaf, with its learned Procrustes transports; the cycle holonomies in Appendix~\ref{app:holonomy_null} probe that twist separately.

\textbf{Hodge harmonic fraction.} On random unit edge cochains (a uniformly random direction for each edge), we compute the fraction of cochain energy in $\ker((\delta^0)^\top)$, the orthogonal complement of $\mathrm{im}(\delta^0)$, and in the harmonic part $\ker((\delta^0)^\top) \cap \ker(\delta^1) \cong H^1$. The two fractions are equal on face-free complexes ($\delta^1 = 0$). For a random cochain we expect the harmonic fraction to be exactly $b_1/|E|$.

\begin{table}[h!]
\centering
\caption{Genuine cohomology and Hodge harmonic fraction across architectures and topologies.}
\label{tab:genuine_h1_hodge}
\small
\begin{tabular}{@{}lrrrrr@{}}
\toprule
\textbf{Topology} & $H^0$ & $H^1$ (algebraic) & $H^2$ & $\ker(\delta^0)^\top$ \textbf{frac.} & \textbf{Harmonic frac.} \\
\midrule
Matching (acyclic) & 180 & 0 & 0 & 0.000 & 0.000 \\
Clique (filled) & 60 & 0 & 240 & 0.600 & 0.000 \\
Relation hypergraph & 6 & \textbf{66} & 0 & 0.184 & \textbf{0.184} \\
\bottomrule
\end{tabular}
\end{table}

The relation hypergraph has expected harmonic fraction exactly $b_1/|E| = 66/360 = 18.33\%$, regardless of architecture; the measured $18.4\%$ is within probe noise. The filled cliques place $60\%$ of their energy outside $\mathrm{im}(\delta^0)$, but all of it is coexact (every cycle is a sum of triangle boundaries), so their harmonic fraction is zero.

\textbf{Six-model LEACE table (whole 60-fact set).} \label{app:leace}

\begin{table}[h!]
\centering
\caption{LEACE vs sheaf $H^0$ on 60-fact retrieval (20 dimensions, top-1 against all 60 facts). LEACE is fitted on paraphrase versus non-paraphrase fact prompts, and sheaf $H^0$ uses the joint-PCA restriction map.}
\label{tab:leace_multi}
\small
\begin{tabular}{@{}lcccccc@{}}
\toprule
\textbf{Model} & \textbf{Full} & $H^0$ (20D) & \textbf{LEACE preserved} & \textbf{LEACE erased} & \textbf{PCA-20} & \textbf{Random-20} \\
\midrule
GPT-2 & 85.3\% & \textbf{95.7}\% & 85.0\% & 2.0\% & 22.7\% & 53.7\% \\
Phi-2 & 97.3\% & 93.0\% & 97.3\% & 3.3\% & 38.0\% & 64.3\% \\
Qwen-2.5-7B & 73.3\% & \textbf{82.0}\% & 75.0\% & 1.0\% & 30.3\% & 50.0\% \\
OLMo-2-7B & 100\% & 97.0\% & 100\% & 2.3\% & 36.7\% & 77.3\% \\
Mistral-7B & 100\% & 80.7\% & 100\% & 2.0\% & 50.0\% & 71.3\% \\
Gemma-2-9B & 99.3\% & 83.7\% & 99.3\% & 1.3\% & 26.7\% & 68.3\% \\
\bottomrule
\end{tabular}
\end{table}

On the same 300 vectors, Table~\ref{tab:leace_multi} compares sheaf $H^0$ with LEACE. Sheaf $H^0$ outperforms LEACE preserved by up to 10.7pp on GPT-2 and Qwen-2.5-7B and outperforms full-space on those models while only using 20 of 768--3584 dimensions. LEACE makes the smallest least-squares edit that makes a binary concept (paraphrase vs non-paraphrase) linearly unpredictable; we instead identify the directions of minimal within-pair variation. LEACE-erased correctly destroys information (1--3\%), which validates the comparison.

\subsection{Simplicial Complex Variants and Scale-Up}
\label{app:simplicial_scaleup}

In the main text we use 1-skeletons and edge dim $k = 8$. We test whether these numbers depend on either choice. Here we fill the multi-layer grid into a 2-complex, fill in the paraphrase cliques of the relation hypergraph, and then recompute Table~\ref{tab:multilayer} for $k = 64$ and $k = 128$.

\paragraph{Triangulated multi-layer grid.}
In Section~\ref{sec:multilayer} we show that filling in each square with two 2-simplices will cause $b_1$ and the harmonic mass to be zero. Here we compute these numbers on the resulting 2-complex.

Each diagonal edge from $(a, l)$ to $(b, l{+}1)$ will have $\omega(e) = (P_{l+1}^\top h_b^{l+1} - \mu_{l+1}) - R_{l \to l+1}^\top (P_l^\top h_a^l - \mu_l)$, which is the vertical-edge formula applied across that pair. The architecture-dependent quantity that remains is the \emph{coexact fraction}. This is the energy of the orthogonal projection of $\omega$ onto $\mathrm{im}\,(\delta^1)^\top$, which is determined by the curl $\delta^1 \omega$. This is the model-implied 2-cochain over the filled-in triangles, or equivalently the curl (circulation) of $\omega$ around each triangle we filled in, i.e.\ the mismatch between the diagonal-then-horizontal and vertical-then-diagonal path sums.

\begin{table}[h!]
\centering
\caption{Triangulated multi-layer grid (each square filled with two triangles sharing a diagonal). Edge dim $k = 64$, $30$ MRPC paraphrase pairs. $b_1 = b_2 = 0$ for every architecture and every $L$, so $\omega$ has no harmonic component (Hodge harmonic mass collapses to machine precision). Reported quantities: model-derived $\omega$ norm squared, coexact fraction (energy explained by the 2-skeleton), and face-cochain norm squared $\|\delta^1 \omega\|^2$. Mistral-7B and Llama-3-8B place under $1\%$ on the 2-skeleton and Qwen-2.5-7B under $3\%$, while GPT-2, Phi-2 and Llama-2-7B place 5--20\%.}
\label{tab:multilayer_triangulated}
\small
\begin{tabular}{@{}lcccccc@{}}
\toprule
& \multicolumn{2}{c}{$L = 2$} & \multicolumn{2}{c}{$L = 4$} & \multicolumn{2}{c}{$L = 8$} \\
\cmidrule(l){2-3}\cmidrule(l){4-5}\cmidrule(l){6-7}
\textbf{Architecture} & coexact & $\|\delta^1\omega\|^2$ & coexact & $\|\delta^1\omega\|^2$ & coexact & $\|\delta^1\omega\|^2$ \\
\midrule
Mistral-7B    & $0.0000$ & $1.4\!\times\!10^{2}$ & $0.0002$ & $8.6\!\times\!10^{2}$ & $0.0006$ & $2.5\!\times\!10^{3}$ \\
Llama-3-8B    & $0.0005$ & $4.7\!\times\!10^{2}$ & $0.0026$ & $1.8\!\times\!10^{3}$ & $0.0075$ & $5.1\!\times\!10^{3}$ \\
Qwen-2.5-7B   & $0.0003$ & $9.8\!\times\!10^{2}$ & $0.0255$ & $6.0\!\times\!10^{5}$ & $0.0247$ & $9.7\!\times\!10^{5}$ \\
GPT-2 (124M)  & $0.0544$ & $6.7\!\times\!10^{3}$ & $0.1219$ & $1.4\!\times\!10^{5}$ & $0.1014$ & $2.2\!\times\!10^{5}$ \\
Phi-2         & $0.0572$ & $5.1\!\times\!10^{3}$ & $0.1895$ & $7.9\!\times\!10^{4}$ & $0.1970$ & $2.0\!\times\!10^{5}$ \\
Llama-2-7B    & $0.0759$ & $6.4\!\times\!10^{4}$ & $0.0815$ & $8.6\!\times\!10^{4}$ & $0.0623$ & $1.3\!\times\!10^{5}$ \\
\bottomrule
\end{tabular}
\end{table}

The coexact fraction in Table~\ref{tab:multilayer_triangulated} recovers the partitioning from Table~\ref{tab:multilayer} for all values of $L \le 8$. The three robust architectures are less than $3\%$ while the three fragile ones are between $5\%$ and $20\%$. So we see the same information that was captured by the harmonic mass of the 1-skeleton is now captured by the coexact direction when we fill the grid into a 2-complex. The layer transitions still do not commute with paraphrase equivalence on the same set of architectures, but this failure is absorbed by the 2-skeleton as a coexact component instead of forming a harmonic class.

\paragraph{Triangulated relation hypergraph.}
By filling in all $\binom{5}{3} = 10$ triangles for each fact's paraphrase 5-clique we have made each fact simply connected, so the 60 within-fact 1-cycles become 1-boundaries. So we now have $b_0 = 6$ (one component per relation), $b_1 = 6$ (these are the only cycles that go through multiple facts, i.e.\ the relation cycles) and $b_2 = 240$ (each $K_5$ 2-skeleton has $b_2 = 4$ and there are $60$ facts). We find that random unit edge cochains have a Hodge harmonic mass of $0.93\%$ on this 2-complex (expectation of $6/660 = 0.91\%$). This is lower than the $18.4\%$ we found on the 1-skeleton ring (Table~\ref{tab:genuine_h1_hodge}). By construction the Hodge mass is a function of the topology and $k = 64$ stalk dim and will be the same for all architectures.

There are $6$ harmonic 1-forms left, which correspond to the relation cycles. There are $b_2 = 240$ new harmonic 2-forms that do not pull back to anything we can interpret on the residual stream. Under the restriction maps we used with shared PCA, the model-implied 2-cochain $\delta^1 \omega$ is 0 to machine precision (all transports are the identity map, so all cochains are exact w.r.t.\ the node potential $P^\top h_v$). So with these restriction maps the 2-complex carries no degree-2 cohomological obstruction beyond what the 1-skeleton already encodes. If we use per-node restriction maps and learn the Procrustes transports we may see different results. We leave this heterogeneous-restriction $H^2$ probe to future work.

\paragraph{Edge-dimension scale-up.}
The body Table~\ref{tab:multilayer} uses $k = 8$. We check whether the differences between architectures, their order by Hodge mass, and the body numbers persist at larger $k$. We repeat the multi-layer grid construction at $k = 64$ ($N = 30$) and $k = 128$ ($N = 80$, as $k$ is bounded by $\min(2N, k_\text{requested})$).

\begin{table}[h!]
\centering
\caption{Multi-layer matching grid Hodge harmonic mass over edge dimension. The body Table~\ref{tab:multilayer} shows $k = 8$. As we increase $k$, the 3 robust architectures stay below the 3 fragile ones for all $L \le 8$, and the range widens to over 3 orders of magnitude. The $k = 128$ row uses $N = 80$ paraphrase pairs to allow for the requested $k$ (otherwise $k$ is capped at $\min(2N, k_\text{req}) = 60$).}
\label{tab:multilayer_scaleup}
\small
\begin{tabular}{@{}lcccccc@{}}
\toprule
& \multicolumn{3}{c}{$k = 64$ ($N=30$)} & \multicolumn{3}{c}{$k = 128$ ($N=80$)} \\
\cmidrule(l){2-4}\cmidrule(l){5-7}
\textbf{Architecture} & $L\!=\!2$ & $L\!=\!4$ & $L\!=\!8$ & $L\!=\!2$ & $L\!=\!4$ & $L\!=\!8$ \\
\midrule
Mistral-7B    & $0.00003$ & $0.0002$ & $0.0005$ & $0.00002$ & $0.0002$ & $0.0005$ \\
Llama-3-8B    & $0.0005$  & $0.0022$ & $0.0061$ & $0.0004$  & $0.0017$ & $0.0049$ \\
Qwen-2.5-7B   & $0.0003$  & $0.0234$ & $0.0248$ & $0.0003$  & $0.0191$ & $0.0215$ \\
GPT-2 (124M)  & $0.050$   & $0.112$  & $0.101$  & $0.041$   & $0.081$  & $0.084$ \\
Phi-2         & $0.054$   & $0.180$  & $0.197$  & $0.053$   & $0.156$  & $0.180$ \\
Llama-2-7B    & $0.067$ & $0.077$ & $0.065$ & $0.083$ & $0.093$ & $0.078$ \\
\bottomrule
\end{tabular}
\end{table}

For all $L \le 8$ and $k$ we see that the robust set of models (Mistral-7B, Llama-3-8B, Qwen-2.5-7B) remains lower than the fragile set of models (GPT-2, Phi-2, Llama-2-7B). Mistral-7B is always the lowest (equal to Qwen-2.5-7B at $k = 8$, $L = 2$). The span widens to over 3 orders of magnitude ($\approx 2 \times 10^{-5}$ to $8 \times 10^{-2}$). Within the fragile set, the order changes depending on $k$ and $L$: Llama-2-7B is the highest at $L = 2$ and Phi-2 is the highest at $L \geq 4$. The body $L = 2$ values are consistent at all edge dimensions ($\approx 10^{-4}$ for Mistral-7B and Qwen-2.5-7B, $\approx 10^{-3}$ for Llama-3-8B and $\approx 0.08$ for Llama-2-7B).

\subsection{Harmonic Mass vs Steering Fragility (Per-Model)}
\label{app:fragility}

Table~\ref{tab:multilayer} shows the two quantities we use for our test in \S\ref{sec:fragility}. Multi-layer Hodge harmonic mass ($L=2$) and steering fragility. Here fragility is fact preservation on CounterFact when steered in a random direction with norm equal to $0.3$ times the hidden-state norm at layer $\lfloor 5L/8 \rfloor$ (lower is more fragile), with $n = 1000$ prompts ($200$ for Phi-2), using the same metric as Table~\ref{tab:steering_four_arch}. Note these quantities are derived from disjoint measurements (MRPC between layers and CounterFact steering). The rank correlation is $\rho = 0.90$ ($p = 0.014$, exact one-sided permutation test over $720$ rankings). The only place the ranks disagree is Qwen-2.5-7B and Llama-3-8B, which is within sampling error ($34.3\%$ vs.\ $36.0\%$ with $n = 1000$).

\subsection{Geometric Picture: Where $H^1$ Lives}
\label{app:geometric_picture}

\begin{figure}[h]
\centering
\includegraphics[width=0.78\linewidth]{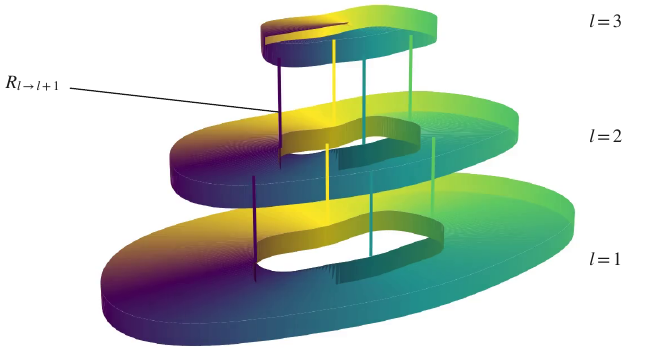}
\caption{Multi-layer matching sheaf schematic, a frame from an animation. Each slab represents a layer $l$, with struts connecting the same sentence between adjacent layers via the Procrustes transport $R_{l \to l+1}$. Traveling up one strut, across layer $l+1$ and down another strut closes an inter-layer square. There are $b_1 = N(L-1)$ such independent squares, and they carry the harmonic mass. Slab shapes are arbitrary and do not represent any topology.}
\label{fig:appendix_manifold}
\end{figure}

\subsection{Proof of the Reduction and Spectral Validation}
\label{app:cohom_validation}

\paragraph{Proof of Proposition~\ref{prop:laplacian_reduction}.} On the matching graph with a single shared $P$ and identity transports, each edge cochain is $\Delta_i^{(k)} = P^\top h_i^{(1)} - P^\top h_i^{(2)} = P^\top \Delta_i$ with $\Delta_i = h_i^{(1)} - h_i^{(2)}$. Hence
\[
L_\mathcal{F} = \frac1N\sum_{i} \Delta_i^{(k)} (\Delta_i^{(k)})^\top = P^\top \Big(\frac1N\sum_i \Delta_i \Delta_i^\top\Big) P = P^\top C_W P,
\]
which is (i). For unit $v$, $v^\top L_\mathcal{F} v = \frac1N\sum_i (v^\top P^\top \Delta_i)^2 $, which is (iii). The Fisher baseline replaces the standard eigenproblem with the generalized one $L_\mathcal{F} v = \lambda (P^\top C_T P) v$, which is (ii). $\qed$ For Remark~\ref{rem:pernode}: with per-node maps the edge cochain is $c_e = P_v^\top h_v - Q_{uv} P_u^\top h_u$, and $\frac1N \sum_e c_e c_e^\top$ contains the cross-terms $P_v^\top h_v h_u^\top P_u Q_{uv}^\top$, which do not factor through a single $P$.

\paragraph{Numerical sanity check.} Using exact arithmetic, the dimension of the kernel of the assembled Laplacian $(\delta^0)^\top \delta^0$ over $C^0$ is $\dim \ker = \dim C^0 - \mathrm{rank}(\delta^0)$. We find that the eigenvalue-threshold estimate matches the rank calculation, up to one eigenvalue near the cutoff, on GPT-2 using 50 MRPC pairs (edge dim $k = 49$, using the $\min(64, d, n-1)$ cap in our script):

\begin{itemize}
\item Spectral $H^0$ dimension (eigenvalues $< 10^{-6}$): 49
\item Algebraic $H^0$ dimension ($\dim C^0 - \mathrm{rank}(\delta^0)$ on the assembled coboundary matrix): 50
\item Coboundary test: $\|\delta^0 \Pi_{H^0} s\| < 10^{-6}$ for all features tested
\end{itemize}

The only discrepancy is one eigenvalue near the cutoff ($\lambda \approx 4 \times 10^{-6} > 10^{-6}$).

\section{Additional Experiments}
\label{app:experiments}

\subsection{Cross-Dataset Generalization}
\label{app:crossval}

We test cross-domain $P$ for $H^0$ retrieval on the held-out CounterFact set (Table~\ref{tab:crossdataset_p}): we train the joint-PCA $P$ on MRPC, PAWS or QQP paraphrases ($n=300$) instead of CounterFact training facts, and apply it to the same CounterFact test set of centroids and queries. Across all 8 models in-domain $P$ averages $81.6\%$, cross-domain $P$ averages $55$--$60\%$, PCA-20 (no kernel) is $30.5\%$ and Random-20 is $51.2\%$. Cross-domain $P$ still benefits from kernel selection vs.\ no kernel ($+5$ to $+30$\,pp) but lags behind in-domain $P$ by $\sim$$25$\,pp. There is some domain dependence in $P$. News (MRPC), adversarial (PAWS) and question (QQP) paraphrases all show non-trivial transfer, but a task that distinguishes facts benefits from $P$ trained on facts. This is a limitation, and we recommend training $P$ in-domain when the downstream retrieval domain is fixed.

\begin{table}[h]
\centering
\caption{Cross-dataset restriction-map transfer (held-out CounterFact, hard same-relation retrieval, $k=20$). $H^0$ is fit with $P$ from the source on the left and evaluated on the same CounterFact test set. Out-of-domain $P$ retains $+5$ to $+30$\,pp over no-kernel baselines but loses $\sim$$25$\,pp relative to in-domain $P$.}
\label{tab:crossdataset_p}
\small
\begin{tabular}{@{}lcccccc@{}}
\toprule
\textbf{Model} & $H^0$ \textbf{CF (in-domain)} & $H^0$ \textbf{MRPC} & $H^0$ \textbf{PAWS} & $H^0$ \textbf{QQP} & \textbf{PCA-20} & \textbf{Random-20} \\
\midrule
GPT-2          & 85.3 & 55.1 & 63.9 & 59.1 & 24.6 & 39.1 \\
Phi-2          & 87.1 & 57.3 & 61.6 & 57.6 & 37.2 & 55.1 \\
Mistral-7B     & 87.1 & 57.8 & 65.9 & 61.2 & 34.5 & 59.2 \\
OLMo-2-7B      & 82.8 & 64.1 & 60.5 & 57.6 & 35.7 & 62.3 \\
Qwen-2.5-7B    & 64.1 & 47.0 & 53.3 & 56.7 & 37.5 & 43.7 \\
Gemma-2-9B     & 77.9 & 48.3 & 47.9 & 51.0 & 27.5 & 43.5 \\
Llama-2-7B     & 80.4 & 60.5 & 60.0 & 53.5 & 22.3 & 50.7 \\
Llama-3-8B     & 88.0 & 56.9 & 65.5 & 60.5 & 24.4 & 55.9 \\
\midrule
\textbf{Mean}  & \textbf{81.6} & 55.9 & 59.8 & 57.1 & 30.5 & 51.2 \\
\bottomrule
\end{tabular}
\end{table}

\subsection{Steering Magnitude Sweep}
\label{app:scale_sweep}

We vary $\alpha \in \{0.05, 0.10, 0.15, 0.20, 0.30, 0.50\}$ on Mistral-7B and Llama-3-8B. With $\alpha$ in the middle range (0.10--0.15) the sheaf-derived methods outperform random steering by 1.0--2.0pp on both models. When we increase the magnitude ($\alpha \geq 0.30$), the concentrated $H^1$ influence over-perturbs the robust models and random becomes comparable. This leads to our recommendations in Appendix~\ref{app:practitioner}.

\subsection{Cycle-Aware $H^0$: Multi-Paraphrase Rings}
\label{app:cycle_h0}

In the body retrieval we use the pair-only graph: each fact has its paraphrases chained by $K{-}1$ edges, giving a forest with $b_1 = 0$. We want to see if connecting all paraphrases of a fact improves $H^0$ retrieval, and if closing a cycle helps as well.

\paragraph{Construction.}
For each fact with $K \in [4, 7]$ paraphrases (we cap at 5) we create a \emph{multi-paraphrase ring}, a closed cycle through all $K$ paraphrases of the fact. There are $N_\text{train} = 350$ facts and thus $\sum_i K_i$ edges and $b_1 = N_\text{train}$ cycles ($\sim$$340$ cycles). The kernel of $L_{\mathcal F}$ now requires invariance across all paraphrases of a fact, which is more restrictive than requiring invariance between pairs. We use the same node activations and joint-PCA $P$ as before. Only the edges differ, and therefore the cycles.

\begin{table}[h]
\centering
\caption{Cycle-aware $H^0$ for hard-CounterFact retrieval. Both
multi-paraphrase rings ($b_1 = N$) and full paraphrase cliques
($b_1 = N(K{-}1)(K{-}2)/2$) beat the pair-only acyclic forest ($b_1 = 0$)
on all eight architectures, mean $+6.5$\,pp. Ring and clique perform within
$0.1$\,pp of each other.}
\label{tab:cycle_h0}
\small
\begin{tabular}{@{}lccccc@{}}
\toprule
\textbf{Model} & $H^0$ \textbf{pair} ($b_1{=}0$) & $H^0$ \textbf{ring} ($b_1{=}N$) & $H^0$ \textbf{clique} ($b_1{\gg}N$) & $\Delta_{\text{ring}}$ & $\Delta_{\text{clique}}$ \\
\midrule
GPT-2          & 87.4 & 90.7 & 89.2 & $+3.4$ & $+1.8$ \\
Phi-2          & 83.1 & 89.6 & 90.5 & $+6.5$ & $+7.4$ \\
Mistral-7B     & 86.7 & 89.8 & 91.4 & $+3.2$ & $+4.7$ \\
OLMo-2-7B      & 84.4 & 91.4 & 91.4 & $+7.0$ & $+7.0$ \\
Qwen-2.5-7B    & 62.3 & 71.8 & 70.7 & $+9.5$ & $+8.4$ \\
Gemma-2-9B     & 75.2 & 84.0 & 84.7 & $+8.8$ & $+9.5$ \\
Llama-2-7B     & 80.8 & 88.9 & 89.6 & $+8.1$ & $+8.8$ \\
Llama-3-8B     & 86.5 & 92.3 & 90.7 & $+5.9$ & $+4.3$ \\
\midrule
\textbf{Mean}  & \textbf{80.8} & \textbf{87.3} & \textbf{87.3} & $\mathbf{+6.5}$ & $\mathbf{+6.5}$ \\
\bottomrule
\end{tabular}
\end{table}

Ring and clique both outperform pair-only on all architectures, by an average of $+6.5$\,pp. Ring and clique perform within $0.1$\,pp of each other on average, so adding more edges than the ring does not help.

\paragraph{Edge-count-matched control.} We want to disentangle the effect of jointly constraining all paraphrases of a fact from closing the cycle. On all facts with at least 6 paraphrases, we compare a ring over 5 nodes ($5$ edges, $b_1 = 1$) with an acyclic path over six paraphrases of the same fact ($5$ edges, $b_1 = 0$). Paired bootstrap confidence intervals on held-out facts show that the path beats pairs by $+7.7$\,pp on average across the 8 models. The ring beats the path by $+0.5$\,pp on average, and the 95\% CI contains 0 for every one of the 8 models. So the gain comes from jointly constraining all paraphrases of a fact. The difference which closes the cycle is just the sum of the differences along the path, and it puts no extra constraint on exact global sections.

\subsection{Restriction Map Sensitivity}
\label{app:restriction_maps}

\begin{figure}[h]
\centering
\includegraphics[width=0.85\linewidth]{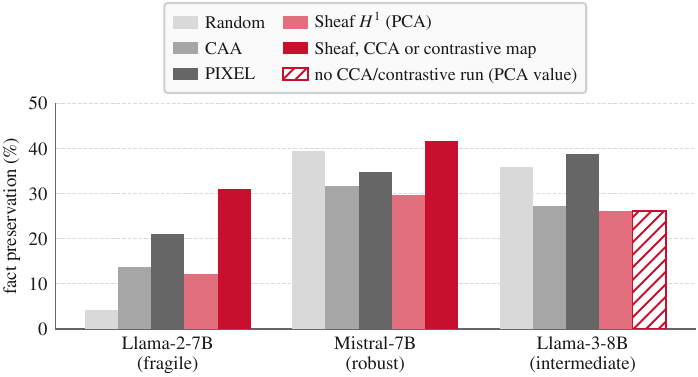}
\caption{Steering on CounterFact at $\alpha=0.3$ ($n=1000$ per model) for the three architectures with a PIXEL run. The hatched bar on Llama-3-8B repeats the PCA $H^1$ value, since that model has no CCA or contrastive run. On fragile Llama-2-7B, random steering collapses to $4.2\%$ and the contrastive sheaf leads at $31.0\%$. On robust Mistral-7B, the CCA sheaf and random tie ($41.6\%$ vs.\ $39.4\%$). On Llama-3-8B, PIXEL and random lead. Same data as Table~\ref{tab:steering_four_arch} and Appendix~\ref{app:pixel_comparison}.}
\label{fig:steering_methods}
\end{figure}

We look at three different restriction maps: PCA (no parameters), CCA (maximize correlation) and contrastive (Fisher discriminant). We find that the best restriction map varies by architecture. CCA improves Mistral-7B by +2.2pp ($n{=}1000$, $\alpha{=}0.3$), while contrastive achieves 31.0\% vs.\ 4.2\% for random on Llama-2-7B (7.4$\times$ increase, $p < 10^{-70}$). Figure~\ref{fig:steering_methods} shows results for 3 of the 4 architectures we evaluated on CounterFact (Llama-2-13B's smaller method grid appears only in Table~\ref{tab:steering_four_arch}). The figure compares Random, CAA, PIXEL, sheaf $H^1$ and the best sheaf method for that architecture. The architecture-dependent pattern is visible across rows: sheaf wins outright on fragile models and is statistically tied with random on robust models.

\paragraph{Restriction-map ablation: PCA vs random vs identity.}
Table~\ref{tab:restriction_ablation} contains the ablation study on restriction maps discussed in \S\ref{sec:leace_comparison}. We calculate $H^0$ using the same activations but with 3 methods: PCA-$P$ (the current method, taking the top 128 PCs from joint-PCA), a random orthonormal $P$ (5 seeds) and identity $P$ (no projection; we use the bottom 20 eigenvectors of $C_W$ in the full $d$-dimensional residual stream). We use hard-CounterFact retrieval and report top-1 accuracy with same-relation distractors, on $n_{\text{test}} = 150$ facts.

\begin{table}[h]
\centering
\caption{Restriction-map ablation (8 architectures, hard-CounterFact retrieval). $H^0$ from $\ker L_{\mathcal F}$ recovers content directions whether $P$ is joint-PCA, a random orthonormal projection, or absent.}
\label{tab:restriction_ablation}
\small
\begin{tabular}{@{}lccccc@{}}
\toprule
\textbf{Model} & $H^0$ \textbf{PCA-}$P$ & $H^0$ \textbf{random-}$P$ & $H^0$ \textbf{identity-}$P$ & \textbf{PCA-20} (no kernel) & \textbf{Random-20} \\
\midrule
GPT-2          & 85.3 & $77.1 \pm 0.8$ & 79.0 & 24.6 & 39.1 \\
Phi-2          & 87.1 & $73.0 \pm 2.0$ & 78.8 & 37.2 & 55.1 \\
Mistral-7B     & 87.1 & $80.6 \pm 1.6$ & 83.3 & 34.5 & 59.2 \\
OLMo-2-7B      & 82.8 & $76.4 \pm 1.6$ & 79.7 & 35.7 & 62.3 \\
Qwen-2.5-7B    & 64.1 & $61.4 \pm 1.3$ & 62.3 & 37.5 & 43.7 \\
Gemma-2-9B     & 77.9 & $69.2 \pm 1.5$ & 76.5 & 27.5 & 43.5 \\
Llama-2-7B     & 80.4 & $76.3 \pm 2.7$ & 80.1 & 22.3 & 50.7 \\
Llama-3-8B     & 88.0 & $80.0 \pm 1.4$ & 79.9 & 24.4 & 55.9 \\
\midrule
\textbf{Mean}  & \textbf{81.6} & \textbf{74.2} & \textbf{77.5} & 30.5 & 51.2 \\
\bottomrule
\end{tabular}
\end{table}

Identity-$P$ (no data-dependent projection) lands within $4.1$\,pp of PCA-$P$ on average, and beats PCA-without-kernel by $+47$\,pp. Random-$P$ averaged over 5 seeds lands within $8$\,pp of PCA-$P$. The kernel selection recovers the content, and the joint-PCA $P$ shrinks the eigenproblem from $\mathbb{R}^d$ to $\mathbb{R}^{128}$. On its own, $P$ does not select content directions.

\subsection{Retrieval Baseline Comparison}
\label{app:baselines}

Results for sheaf $H^0$ compared to various subspace and representation learning approaches (LDA, CCA, PLS, top-20 PCA and NCA metric learning), applied to the same problem, are shown in Table~\ref{tab:retrieval_baselines}. As before (Table~\ref{tab:leace_held_out}), we test on held-out hard-CounterFact examples, using identical activations, a 70/30 split of facts, a $20$-dimensional subspace, and top-1 evaluation. Except for W-PCA, all results are from a single common run. Values for sheaf $H^0$ and Full are all within $1$\,pp of those in Table~\ref{tab:leace_held_out}, which used a different random seed and different sample of examples. For W-PCA (whitened PCA), we first mean-subtract all activations, and multiply each of the top-20 eigenvectors by $1/\sqrt{\lambda}$. We ran this separately, but otherwise with identical protocol to the above. This gives an improvement of up to $17$\,pp over plain PCA-20. However, on every architecture W-PCA is still $27$ to $48$\,pp worse than sheaf $H^0$. As described above, LDA and NCA use the label indicating which fact is present. All other methods: sheaf $H^0$, CCA, PLS, PCA-20 and W-PCA, are unlabeled. On all 8 models sheaf $H^0$ exceeds every one of these by $+9$ to $+36$\,pp. On average over models it is within $\sim$$2$\,pp of the best labeled method, although this is as much as $8$\,pp for Qwen, and it is best overall for Phi-2, Gemma-2 and Llama-3. We show some standard supervised models for reference only: SimCSE achieves $97.5\%$ and a sentence transformer $99.1\%$, both trained on very large numbers of paraphrases. Note that as a by-product of the sheaf operator, we also obtain the $H^1$ influence subspace, the steering decomposition, and the harmonic mass measure of fragility. None of these can be obtained from either LDA or NCA.

\begin{table}[h]
\centering
\caption{Retrieval baseline comparison (held-out hard-CounterFact, top-1 \%, $k = 20$, same protocol as Table~\ref{tab:leace_held_out}). Label-free: sheaf $H^0$, CCA, PLS, PCA-20 (top-20 principal components without kernel selection, as in Table~\ref{tab:leace_held_out}), W-PCA (the same 20 components whitened, each scaled by $1/\sqrt{\lambda}$ on centred activations). Label-requiring: LDA, NCA. Sheaf $H^0$ bold where it beats both LDA and NCA.}
\label{tab:retrieval_baselines}
\small
\begin{tabular}{@{}lcccccccc@{}}
\toprule
\textbf{Model} & \textbf{Sheaf} $H^0$ & \textbf{LDA} & \textbf{NCA} & \textbf{CCA} & \textbf{PLS} & \textbf{PCA-20} & \textbf{W-PCA} & \textbf{Full} \\
\midrule
GPT-2        & 84.9 & 87.4 & 88.3 & 47.9 & 49.0 & 24.4 & 39.5 & 43.1\\
Phi-2        & \textbf{87.6} & 84.9 & 83.3 & 58.9 & 64.1 & 37.9 & 42.7 & 69.1\\
Mistral-7B   & 86.9 & 90.1 & 88.3 & 60.7 & 63.7 & 34.3 & 45.1 & 77.4\\
OLMo-2-7B    & 82.6 & 88.5 & 86.9 & 62.8 & 67.5 & 36.6 & 37.2 & 81.3\\
Qwen-2.5-7B  & 64.8 & 72.5 & 72.9 & 50.8 & 56.0 & 36.6 & 38.1 & 50.3\\
Gemma-2-9B   & \textbf{77.4} & 76.1 & 76.3 & 51.2 & 50.1 & 27.3 & 37.2 & 54.4\\
Llama-2-7B   & 80.6 & 83.5 & 84.4 & 49.7 & 49.9 & 21.9 & 39.3 & 68.2\\
Llama-3-8B   & \textbf{87.8} & 86.5 & 86.0 & 53.3 & 55.5 & 24.4 & 39.3 & 72.7\\
\midrule
\textbf{Mean} & 81.6 & 83.7 & 83.3 & 54.4 & 57.0 & 30.4 & 39.8 & 64.6\\
\bottomrule
\end{tabular}
\end{table}

\subsection{Sheaf Eigenbasis Ablation}
\label{app:sheaf_basis}

This ablation uses the eigenbasis of $L_\mathcal{F}$ directly. At every other layer, $H^1$ is the top $20$ eigenvectors of $L_\mathcal{F}$ with the joint-PCA restriction map ($k = 128$), fit on mean-pooled block outputs of $200$ MRPC paraphrase pairs, and $H^0$ is the bottom $20$. On $100$ held-out MRPC sentences, the block output is projected onto the orthogonal complement of the subspace at every position, the rest of the forward pass runs, and the effect is the KL divergence of the next-token distribution at the final position. The control is $20$ Haar-random $20$-dimensional subspaces, which are not matched to $H^1$ in activation energy; Table~\ref{tab:causal_nine_arch} is the variability-matched comparison. Table~\ref{tab:sheaf_basis} reports, per model, the ratio of mean KL ($H^1$ over random) as the median and range over layers, the smallest fraction of prompts over layers for which $H^1$ exceeds all $20$ random draws, and the median $H^0$ ratio.

\begin{table}[h]
\centering
\caption{Sheaf eigenbasis ablation: ratio of mean next-token KL from projecting out $H^1$ (top $20$ eigenvectors of $L_\mathcal{F}$) to that from $20$ random $20$-dimensional subspaces at every other layer. Median and range are over layers. The prompts column is the smallest fraction over layers of held-out prompts on which $H^1$ exceeds all $20$ draws.}
\label{tab:sheaf_basis}
\small
\begin{tabular}{@{}lccccc@{}}
\toprule
\textbf{Model} & \textbf{Layers} & $H^1$\textbf{/random, median} & \textbf{Range} & \textbf{Prompts} & $H^0$\textbf{/random, median} \\
\midrule
GPT-2 (124M) & 6 & $17.8\times$ & 13.7--53.5 & 99\% & $0.19$ \\
Phi-2 & 16 & $84.1\times$ & 18.4--122 & 97\% & $0.56$ \\
Mistral-7B & 16 & $398\times$ & 70.0--3075 & 100\% & $0.54$ \\
Qwen-2.5-7B & 14 & $127\times$ & 78.7--293 & 99\% & $0.45$ \\
OLMo-2-7B & 16 & $182\times$ & 112--1048 & 100\% & $2.21$ \\
Gemma-2-9B & 21 & $240\times$ & 75.8--1211 & 100\% & $0.74$ \\
Llama-2-7B & 16 & $287\times$ & 102--2283 & 100\% & $0.32$ \\
Llama-3-8B & 16 & $176\times$ & 75.7--409 & 100\% & $0.29$ \\
Llama-2-13B & 20 & $327\times$ & 191--645 & 100\% & $0.47$ \\
\bottomrule
\end{tabular}
\end{table}

\subsection{Dormancy and the Subspace-Patching Illusion}
\label{app:illusion}

Makelov et al.~\cite{makelov2024subspace} analyze an illusion for \emph{interchange} patching, $a + (p_A - p_B)v$ with $v = (v_{\mathrm{disc}} + v_{\mathrm{dormant}})/\sqrt{2}$: activating variation along a causally disconnected direction can turn on a dormant direction. By contrast, Section~\ref{sec:influence} zero-ablates the subspace via projection onto its orthogonal complement. We conduct two checks to rule out a dormant-pathway interpretation of our results. By definition a dormant direction is one that sees no differential activation on-distribution. We compute the on-distribution projection spread of $H^1$, $H^0$, and five variance-matched random subspaces (same dimension, norm, and eigenspectrum). We then ablate each and report fact recall from the greedy 50-token continuation of the bare CounterFact prompt. This prompt does not contain the answer ($n = 100$ for GPT-2 and $150$ for the 7B models). $H^1$ is the least dormant subspace we examine, and ablating it collapses fact recall, while ablating the matched random subspace leaves recall at baseline (Table~\ref{tab:illusion}). GPT-2 only recalls $15\%$ of these facts unaided, so its row is the weakest. Because the generated text itself is modified by this intervention, the effect is behavioural and does not rely on the KL tail. We observe an intervention KL ratio of $15.7\times$ on GPT-2. On the 7B models, the KL saturates numerically (median raw KL of $7.1$ vs.\ $0.003$ nats on Mistral-7B), so here we rely on the collapse of recall as our evidence.

\begin{table}[h]
\centering
\caption{Dormancy and behavioural ablation. Dormancy ratio is the on-distribution projection spread of $H^1$ divided by the mean spread of variance-matched random subspaces. Recall is the fraction of greedy continuations of the bare prompt that contain the target entity. Layer $\ell \approx 2L/3$, $k = 20$, five random draws, seed 42.}
\label{tab:illusion}
\small
\begin{tabular}{@{}lcccc@{}}
\toprule
\textbf{Model} & \textbf{Dormancy ratio} ($H^1$/random) & \textbf{Recall, no ablation} & $H^1$ \textbf{ablated} & \textbf{Random ablated} \\
\midrule
GPT-2      & $2.5\times$ & 15.0\% & 2.0\% & 7.5\% \\
Mistral-7B & $6.6\times$ & 46.7\% & 0.0\% & 47.0\% \\
Llama-2-7B & $5.6\times$ & 47.3\% & 4.7\% & 48.7\% \\
\bottomrule
\end{tabular}
\end{table}

\subsection{Cross-Lingual Invariance Probe}
\label{app:crosslingual}

All that is required is a set of putatively equivalent contexts, together with corresponding model activations. To test generalisation to translation, we take English--French sentence pairs from OPUS-100 to form sheaf edges over Llama-3-8B, layer 21, edge dimension $128$, $k = 20$, seed 42. We learn $H^0$ using $100$ pairs and test retrieval of the correct translation, by top-1 accuracy, over another $100$ unseen pairs. Results are shown in Table~\ref{tab:crosslingual}. The $H^0$ kernel keeps cross-lingual retrieval, performing as well as or better than the original $4096$-d representation for EN$\to$FR, and the $H^1$ complement loses it. We see similar behaviour for paraphrase and translation, although we do not investigate the latter further here.

\begin{table}[h]
\centering
\caption{Held-out cross-lingual retrieval on Llama-3-8B (top-1 \%, $100$ test pairs, $H^0$ fit on $100$ disjoint training pairs).}
\label{tab:crosslingual}
\small
\begin{tabular}{@{}lcc@{}}
\toprule
\textbf{Subspace} & \textbf{FR$\to$EN} & \textbf{EN$\to$FR} \\
\midrule
$H^0$ kernel (20-dim)          & 76 & 78 \\
$H^1$ complement               & 6  & 3  \\
Full hidden state (4096-dim)   & 82 & 59 \\
\bottomrule
\end{tabular}
\end{table}

\subsection{Per-Architecture Layer Profiles}
\label{app:layer_dynamics}

\begin{figure}[h]
\centering
\includegraphics[width=\linewidth]{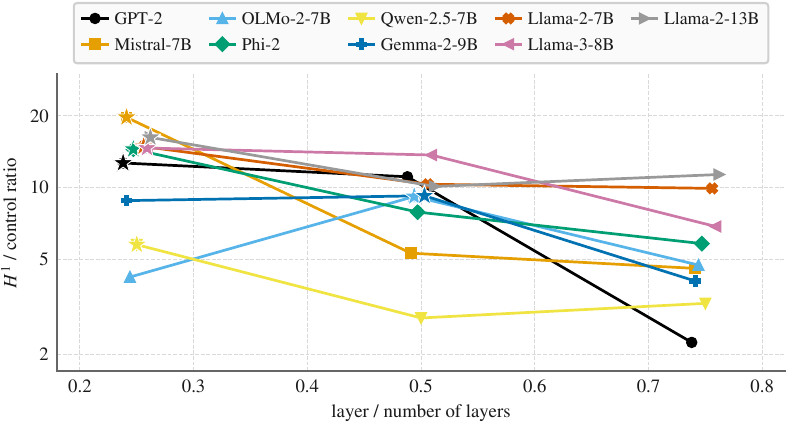}
\caption{$H^1$-to-control effect ratio (protocol $b$ of Table~\ref{tab:causal_nine_arch}) at three depths per architecture, $L/4$, $L/2$ and $3L/4$, plotted against normalized depth (layer index over number of layers). Stars mark each architecture's largest measured ratio, and points are offset slightly in depth so overlapping peaks stay visible. Only measured points are shown. The GPT-2 and Llama values here come from protocol $b$ and differ from their protocol-$a$ rows in Table~\ref{tab:causal_nine_arch}.}
\label{fig:layer_profile}
\end{figure}

In the experiments using 3 different depths (Table~\ref{tab:causal_nine_arch}, protocol $b$), we plot the $H^1$-to-control ratio at $L/4$, $L/2$ and $3L/4$ for each of the nine models in Figure~\ref{fig:layer_profile}. The depth of the maximum varies across models: $L/4$ for GPT-2, Mistral-7B, Llama-3-8B, Llama-2-7B, Llama-2-13B, Phi-2 and Qwen-2.5-7B, and $L/2$ for OLMo-2-7B and Gemma-2-9B. Sampling only 3 depths per model cannot resolve how this changes over all layers. For GPT-2, Llama-2-7B, Llama-3-8B and Llama-2-13B, the table reports the value at $L/2$ from our separate $1000$-sentence run (protocol $a$).

\subsection{McNemar Tests with Multiple Comparisons Correction}
\label{app:significance}

All main comparisons (sheaf vs. random, sheaf vs. Fisher) are evaluated via McNemar's test ($n$=1000 per model) with Benjamini-Hochberg correction for 16 simultaneous tests ($q < 0.05$). 13 of 16 tests survive correction, with the largest effects on Llama-2-7B (Cohen's $h = 2\arcsin\sqrt{p_1} - 2\arcsin\sqrt{p_2} = +0.35$, $p < 0.001$).

\subsection{PIXEL Head-to-Head on CounterFact}
\label{app:pixel_comparison}

PIXEL~\cite{yu2025pixel} (Yu et al., 2025) computes a set of steering directions, one per layer, from two complementary views of the prompt: the average hidden state over the last $k_\text{tail} = \mathrm{clip}(\mathrm{round}(0.10 \cdot \text{plen}), 3, 8)$ prompt tokens, and the hidden state at the end token only. The differences between paired positive and negative samples are stacked for these 2 views, and the top weighted PCA direction ($r{=}1$) is taken, giving the unit vector $u_\ell$ for each layer. Given any (layer, prompt-position) site, the corresponding magnitude is then calculated at inference in closed form:
\begin{align*}
a &= h^\top u, \quad B = \sqrt{\|h\|^2 - a^2}, \quad T = B s / \sqrt{1 - s^2}, \\
\alpha &= \mathrm{clip}\big(\max(\max(0, -a/c),\; (T - a)/c),\;\;[\alpha_{\min}, \alpha_{\max}]\big),
\end{align*}
where $s = z_\text{target}$ is the target z-score (we use the PIXEL default $s = 0.85$) and $c$ is the original direction norm (here $c = 1$).

We follow the published procedure to obtain PIXEL's directions and per-position magnitudes. However, we use a single global direction, extracted once from the casual/formal CAA pairs, rather than a separate direction for each question: CounterFact has no per-question quantity to play the role of the MC2 score that PIXEL uses for gating on TruthfulQA. We only inject at the last $k_\text{tail}$ positions when passing the prompt through the network. As in \S\ref{sec:steering_diagnostic}, we use a repetition penalty of $1.3$, greedy decoding of up to $100$ new tokens, layer $\ell = \lfloor 5L/8 \rfloor = 20$, and $4$-bit quantization. Hence PIXEL differs from our other methods only in how the direction is extracted and how the per-position magnitudes are calibrated.

We run PIXEL with $n = 1000$ and $\alpha_\text{clip} = [0, 200]$, which leaves the closed form free to choose any magnitude, and report results in Table~\ref{tab:steering_four_arch}. Like our methods, PIXEL is much better than CAA on the fragile Llama-2-7B, and is matched or beaten by random on the robust Mistral-7B. So again the influence-concentration mechanism (\S\ref{sec:steering_diagnostic}) appears to be common to different ways of extracting directions, in line with the architecture-diagnostic reading. It is advantageous to use two views when deriving directions, compared to simply averaging the differences as in CAA. However, calibrating a magnitude for each position in closed form is insufficient to remove the fragility of the architecture. The best single method on Llama-2-7B remains contrastive sheaf steering with the architecture-matched restriction map, at $31.0\%$; on Mistral-7B it is CCA sheaf steering, at $41.6\%$. No sheaf method outperforms random on Llama-3-8B at $\alpha = 0.30$.

What we didn't do: PIXEL applies a per-question soft gate based on a labeled MC2 score on TruthfulQA, but this is not available on CounterFact, hence we use one global direction. Activations may optionally also be altered at continuation positions (``include-pred''); this is substantially more involved, since it means modifying activations during incremental decoding with the KV cache, and goes beyond the direction-vs-direction comparison we want to make. PIXEL also tunes residual\_scale and auto-flip for each benchmark. We keep the defaults ($1.0$ and no flip) and use sign-aligned PCA. All of these variations concern how PIXEL sets magnitudes; we use their recipe for extracting directions, as this is the part we want to compare against our $H^1$ direction.

\subsection{Holonomy Null Calibration}
\label{app:holonomy_null}

The holonomy around a cycle $\gamma$ is the composition $H_\gamma$ of the orthogonal transports along that cycle's edges. The obstruction satisfies $\|I - H_\gamma\|_F^2 = 2k - 2\,\mathrm{tr}\,H_\gamma \le 4k$, so it's bounded by $2\sqrt{k}$, and a Haar-random orthogonal matrix gives $\mathbb{E}\|I - H_\gamma\|_F^2 = 2k$ (RMS of $\sqrt{2k}$), so raw holonomy magnitudes need calibrating with the nulls below.

We compare the cycle-holonomy obstruction $\|I - H_\gamma\|_F$ to two null distributions, and also vary the number of samples to assess the stability of the estimate (full data in Table~\ref{tab:holonomy_null}). In all cases we take edge dim $k=8$, $11$ fundamental cycles, layer $\ell = 8$ for GPT-2 or $\ell = 16$ for all other models, and $6$ token samples at each node.

\textbf{Shuffled-paraphrase null.} With $20$ different random seeds, randomly reassign which paraphrase belongs to which fact \emph{within each relation}, refit all per-node PCA $P_v$ and Procrustes transports $Q_{uv}$, and measure the resulting holonomies. The observed obstructions are all $0.10$--$0.29$ higher than this null, across all four models, and none of the differences is significant (one-sided $p = 0.10$--$0.33$).

\textbf{Bootstrap on tokens-per-node.} Again refit all $P_v$ and $Q_{uv}$, taking $N \in \{2, 3, 4, 5, 6\}$ token samples per node. All four models show lower values of the obstruction for larger $N$, ranging from $3.1$--$4.4$ at $N = 2$ down to $2.2$--$2.8$ at $N = 6$. Hence some of the observed magnitude should be attributed to finite-sample noise in the Procrustes fits, which are underdetermined when $N < k$.

\textbf{Identity-transport sanity.} Setting every $Q_{uv} = I$ gives $\|I - H_\gamma\|_F = 0$ to machine precision, and on synthetic flat data, where every true holonomy is $I$, the pipeline recovers $\|I - H_\gamma\|_F < 10^{-4}$.

\begin{table}[h]
\centering
\caption{Holonomy null calibration on the relation hypergraph. ``Observed'' = real paraphrase assignments. ``Shuffled null'' = mean over $20$ within-relation paraphrase shuffles. $p$ is the one-sided permutation $p$-value for $P(\text{shuffled} \geq \text{observed})$, smaller meaning observed obstruction exceeds null. Bootstrap-$N$ obstructions are mean $\pm$ std over $11$ cycles at the listed token count.}
\label{tab:holonomy_null}
\small
\begin{tabular}{@{}lccccccc@{}}
\toprule
& \textbf{Observed} & \textbf{Shuffled} & $p$ & $N{=}2$ & $N{=}3$ & $N{=}4$ & $N{=}6$ \\
\midrule
GPT-2 (124M) & $2.92$ & $2.78$ & $0.24$ & $3.09$ & $3.15$ & $2.55$ & $2.43$ \\
Mistral-7B   & $2.52$ & $2.42$ & $0.33$ & $3.20$ & $3.31$ & $2.64$ & $2.18$ \\
Qwen-2.5-7B  & $3.13$ & $2.84$ & $0.10$ & $3.82$ & $3.43$ & $2.90$ & $2.78$ \\
Phi-2        & $3.01$ & $2.76$ & $0.24$ & $4.37$ & $3.37$ & $3.28$ & $2.36$ \\
\bottomrule
\end{tabular}
\end{table}

So the cycle holonomy on the relation hypergraph is far from Haar-random, but at this sample size it cannot be told apart from that obtained by randomly reassigning paraphrases within a relation: the transports carry relation structure that survives the reassignment, and we cannot detect any dependence on the particular facts, for any model.

\subsection{Node-Space Operationalizations of Relation-Hypergraph $H^1$}
\label{app:h1_node_space}

From \S\ref{sec:relation_hypergraph}, our relation hypergraph has $\dim H^1 = 66$ (algebraically) and carries $18.4\%$ Hodge harmonic mass. We found no downstream advantage to lifting this to node space.

\textbf{Harmonics as features.} One natural experiment was to project the node activations into the harmonic subspace $\ker(L_1)$, pulled back to the node stalks using the adjoints of $\delta^0$. For the 60-fact problem, with hard negatives sharing the same relation, this achieved only $36.3\%$ retrieval (\S\ref{sec:fact_probe}). This is worse than the random-20 baseline, which achieves $49.1\%$ with GPT-2. It seems reasonable to expect performance against this type of query to be poor, since the harmonics measure failure of local pieces to glue together, not facts. In particular, a direction stable around the cycles will be stable over the whole relation but not within a fact, so the basis does not match the retrieval target.

Nevertheless, the algebraic result holds, as established directly from the spectrum and the Hodge mass in Appendix~\ref{app:algebraic_h1}. There is simply no retrieval gain from projecting relation-hypergraph $H^1$ to node space. There are other approaches we have not tried, for example using the cycles more directly for steering, learning supervised features with the harmonic basis as a starting point, or learning per-node restriction maps optimized for downstream prediction.

\subsection{Sheaf-Regularized SSL Collapse Prevention (Per-Architecture)}
\label{app:collapse}

Results for the individual architectures, as described briefly in \S\ref{sec:jepa}: we use an MLP with 2 hidden layers ($d \to 256 \to 256 \to 64$). We train on $300$ MRPC paraphrase pairs for $1500$ steps with Adam. Optimizing just the naive MSE between pairs, or $\mathrm{tr}(L_\mathcal{F})$ alone, causes the encoder to collapse to an effective rank of $14$--$42$ out of $64$ dimensions, reaching pair MSE indistinguishable from zero. Using $\mathrm{tr}(L_\mathcal{F}) - \beta\log\det \mathrm{Cov}$ gives effective ranks of $57$--$60$ out of $64$, uniformly over all seven architectures.

\begin{table}[h]
\centering
\small
\setlength{\tabcolsep}{3pt}
\begin{tabular}{@{}llcccc@{}}
\toprule
\textbf{Embedding} & \textbf{Loss} & \textbf{Top-1 $\sigma^2$ frac.} & \textbf{Effective rank} & \textbf{Feature var.} & \textbf{Pair MSE} \\
\midrule
\multirow{3}{*}{GPT-2 (768D)}
& Naive MSE                                              & $0.149$ & $16.0 / 64$ & $0.006$    & $0.0000$ \\
& $\mathrm{tr}(L_\mathcal{F})$ only                       & $0.151$ & $16.6 / 64$ & $0.008$    & $0.0000$ \\
& $\mathrm{tr}(L_\mathcal{F}) - \beta\log\det \mathrm{Cov}$    & $0.109$ & $\mathbf{58.5 / 64}$ & $\mathbf{4991}$ & $0.0149$ \\
\midrule
\multirow{3}{*}{Mistral-7B (4096D)}
& Naive MSE                                              & $0.134$ & $34.4 / 64$ & $0.005$    & $0.0000$ \\
& $\mathrm{tr}(L_\mathcal{F})$ only                       & $0.121$ & $33.1 / 64$ & $0.006$    & $0.0000$ \\
& $\mathrm{tr}(L_\mathcal{F}) - \beta\log\det \mathrm{Cov}$    & $0.094$ & $\mathbf{58.0 / 64}$ & $\mathbf{8418}$ & $0.0186$ \\
\midrule
\multirow{3}{*}{Qwen-2.5-7B (3584D)}
& Naive MSE                                              & $0.107$ & $35.9 / 64$ & $0.006$    & $0.0000$ \\
& $\mathrm{tr}(L_\mathcal{F})$ only                       & $0.119$ & $27.5 / 64$ & $0.005$    & $0.0000$ \\
& $\mathrm{tr}(L_\mathcal{F}) - \beta\log\det \mathrm{Cov}$    & $0.143$ & $\mathbf{58.5 / 64}$ & $\mathbf{6213}$ & $0.0202$ \\
\midrule
\multirow{3}{*}{Llama-2-7B (4096D)}
& Naive MSE                                              & $0.155$ & $36.1 / 64$ & $0.006$    & $0.0000$ \\
& $\mathrm{tr}(L_\mathcal{F})$ only                       & $0.115$ & $42.5 / 64$ & $0.006$    & $0.0000$ \\
& $\mathrm{tr}(L_\mathcal{F}) - \beta\log\det \mathrm{Cov}$    & $0.082$ & $\mathbf{59.5 / 64}$ & $\mathbf{6512}$ & $0.0128$ \\
\midrule
\multirow{3}{*}{Llama-3-8B (4096D)}
& Naive MSE                                              & $0.169$ & $22.2 / 64$ & $0.006$    & $0.0000$ \\
& $\mathrm{tr}(L_\mathcal{F})$ only                       & $0.115$ & $33.6 / 64$ & $0.006$    & $0.0000$ \\
& $\mathrm{tr}(L_\mathcal{F}) - \beta\log\det \mathrm{Cov}$    & $0.109$ & $\mathbf{58.0 / 64}$ & $\mathbf{6843}$ & $0.0316$ \\
\midrule
\multirow{3}{*}{OLMo-2-7B (4096D)}
& Naive MSE                                              & $0.168$ & $23.3 / 64$ & $0.006$    & $0.0000$ \\
& $\mathrm{tr}(L_\mathcal{F})$ only                       & $0.420$ & $14.3 / 64$ & $0.005$    & $0.0000$ \\
& $\mathrm{tr}(L_\mathcal{F}) - \beta\log\det \mathrm{Cov}$    & $0.184$ & $\mathbf{57.0 / 64}$ & $\mathbf{4127}$ & $0.0134$ \\
\midrule
\multirow{3}{*}{Gemma-2-9B (3584D)}
& Naive MSE                                              & $0.185$ & $23.4 / 64$ & $0.006$    & $0.0000$ \\
& $\mathrm{tr}(L_\mathcal{F})$ only                       & $0.145$ & $23.0 / 64$ & $0.006$    & $0.0000$ \\
& $\mathrm{tr}(L_\mathcal{F}) - \beta\log\det \mathrm{Cov}$    & $0.104$ & $\mathbf{57.8 / 64}$ & $\mathbf{3686}$ & $0.0133$ \\
\bottomrule
\end{tabular}
\caption{Collapse prevention per architecture: top-1 variance fraction, effective rank (of 64), feature variance and paraphrase-pair MSE under naive MSE, the Laplacian energy $\mathrm{tr}(L_\mathcal{F})$ alone, and the Laplacian energy with a log-determinant covariance term.}
\label{tab:collapse}
\end{table}

\subsection{MLST on STS-B (Full Table)}
\label{app:mlst}

Results on STS-B (Spearman/Pearson) for Multi-Layer Sheaf Consistency Training against a frozen baseline, SimCSE, and VICReg. Encoders are frozen, with edge dim $k = 64$, trainable projections $P_l$ for each layer and transitions $T_{l \to l+1}$, trained for $2000$ steps using Adam over MRPC paraphrase pairs (positive labels only). We use GPT-2 with layers $\{3, 6, 9, 12\}$ and 10 seeds, and Mistral-7B with layers $\{8, 16, 24, 32\}$ and 5 seeds, with identical hyperparameters. As an ablation, \emph{nosheaf} removes the $T$-transitions and the cycle-holonomy term, so it only requires each layer individually to be consistent over paraphrases, together with the variance/covariance regulariser. Read-out modes: \emph{final} = $P_L^\top h^L$; \emph{concat} = $[P_l^\top h^l]_l$ stacked; \emph{section} = concat together with $T$-transported predictions $T_{l-1\to l} P_{l-1}^\top h^{l-1}$.

\begin{table}[h]
\centering
\small
\begin{tabular}{@{}lcc|cc@{}}
\toprule
 & \multicolumn{2}{c|}{\textbf{GPT-2} ($n=10$)} & \multicolumn{2}{c}{\textbf{Mistral-7B} ($n=5$)} \\
\textbf{Method} & \textbf{Spearman} $\uparrow$ & \textbf{Pearson} $\uparrow$ & \textbf{Spearman} $\uparrow$ & \textbf{Pearson} $\uparrow$ \\
\midrule
Vanilla (no training)                  & $0.428$                  & $0.312$ & $0.533$                    & $0.498$ \\
SimCSE (final-layer, InfoNCE)          & $0.578 \pm 0.014$        & $0.567$ & $0.566 \pm 0.016$          & $0.549$ \\
VICReg (final-layer, inv + var + cov)  & $0.623 \pm 0.006$        & $0.602$ & $\mathbf{0.645 \pm 0.025}$ & $\mathbf{0.635}$ \\
MLST nosheaf (final)                   & $0.621 \pm 0.005$        & $0.598$ & $0.634 \pm 0.025$          & $0.626$ \\
\textbf{MLST nosheaf (concat)}         & $\mathbf{0.668 \pm 0.005}$ & $\mathbf{0.650}$ & $0.644 \pm 0.009$  & $0.630$ \\
MLST full (final)                      & $0.620 \pm 0.006$        & $0.597$ & $0.622 \pm 0.062$          & $0.616$ \\
MLST full (concat)                     & $0.658 \pm 0.007$        & $0.635$ & $0.631 \pm 0.033$          & $0.618$ \\
MLST full (section)                    & $0.658 \pm 0.006$        & $0.634$ & $0.629 \pm 0.021$          & $0.611$ \\
\bottomrule
\end{tabular}
\caption{Multi-layer sheaf consistency training (MLST) on STS-B with frozen encoders: Spearman and Pearson correlation, mean $\pm$ s.d. over seeds. \emph{nosheaf} drops the transitions and the cycle term.}
\label{tab:mlst}
\end{table}

Paired-$t$ tests on per-seed Spearman, GPT-2 ($n = 10$): MLST-nosheaf-concat $-$ VICReg $\Delta = +0.046$, $p = 6 \times 10^{-9}$; MLST-section $-$ VICReg $\Delta = +0.035$, $p = 1 \times 10^{-6}$; MLST-section $-$ SimCSE $\Delta = +0.080$, $p = 2 \times 10^{-7}$; MLST-section $-$ MLST-nosheaf-concat $\Delta = -0.010$, $p = 9 \times 10^{-4}$. Mistral-7B ($n = 5$): MLST-section $-$ VICReg $\Delta = -0.016$, $p = 0.20$; MLST-nosheaf-concat $-$ VICReg $\Delta = -0.001$, $p = 0.94$; MLST-section $-$ SimCSE $\Delta = +0.063$, $p = 0.005$. Cross-architecture rank-correlation across the eight trained methods is $\rho = 0.74$, $p = 0.037$. For both architectures, it is beneficial to project multiple layers and apply the variance/covariance regularization, which outperforms the single-layer baselines. There does not appear to be any benefit at this scale to including the sheaf transitions and cycle apparatus as a training signal; if anything it is slightly negative. The cycle-holonomy term is too easily satisfied by aligning the paraphrase means and overlaps with the consistency loss, so this is not surprising, and there is no consistent gradient signal to gain from the $Q$-transitions in addition to simply concatenating the layers. This may not hold for larger encoders, harder tasks, or richer cycle structure, which we leave to future work.

\section{Reproducibility}
\label{app:reproducibility}

\subsection{Hyperparameter Sensitivity}
\label{app:ablation}

The spectral gap (ratio of maximum to minimum eigenvalues) stays in 91.7--95.6$\times$ across edge dimensions $k \in \{32, 64, 128\}$ (the count $m$ selects eigenvectors and does not change the spectrum); the content/context boundary is stable under these choices.

\paragraph{$m$ sensitivity for $H^0$ retrieval.}
We vary $m \in \{5, 10, 20, 40, 80, 128\}$ (with edge dimension $k=128$ fixed) on hard CounterFact retrieval (Table~\ref{tab:m_sensitivity}). For all four models tested there is a plateau over $m \in [20, 80]$, and $m = 40$ consistently performs slightly better than $m = 20$ ($+4$ to $+6$\,pp). For the boundary case $m = 128 = k$ there are no eigenvectors above the spectral gap (no kernel selection), and performance collapses to $36$--$53\%$. Using the full hidden state without projection gives $43$--$78\%$. We set $m = 20$ \emph{a priori}, well away from the $k$ boundary; the results are not tuned on the test set and are stable across the plateau.

\begin{table}[h]
\centering
\caption{$H^0$ subspace dimension $m$ sweep (hard CounterFact retrieval, edge dim $k = 128$). Results are stable across $m \in \{20, 40, 80\}$. At $m = 128 = k$ no kernel is selected and retrieval collapses.}
\label{tab:m_sensitivity}
\small
\begin{tabular}{@{}lccccccc@{}}
\toprule
\textbf{Model} & $m{=}5$ & $m{=}10$ & $m{=}20$ & $m{=}40$ & $m{=}80$ & $m{=}128$ & \textbf{Full} \\
\midrule
GPT-2          & 63.7 & 74.3 & 85.3 & \textbf{91.2} & 87.8 & 35.9 & 43.1 \\
Phi-2          & 67.3 & 75.8 & 87.1 & \textbf{90.3} & 85.3 & 52.8 & 69.1 \\
Mistral-7B     & 70.9 & 82.4 & 87.1 & \textbf{90.3} & 86.7 & 48.5 & 77.7 \\
Llama-2-7B     & 62.8 & 74.0 & 80.4 & \textbf{85.3} & 85.1 & 35.2 & 68.2 \\
\bottomrule
\end{tabular}
\end{table}

\paragraph{Pooling sensitivity.}
We compare three pooling choices (mean over all tokens, last non-pad token, first token) by extracting activations under each and rerunning the full $H^0$ pipeline (Table~\ref{tab:pooling_sensitivity}). Mean pooling is best across all four architectures ($H^0$ at $80$--$87\%$); last-token pooling drops $10$--$30$\,pp; first-token pooling collapses (autoregressive models have very limited content at position 0). Sheaf $H^0$ beats the matched Full hidden state under mean and last-token pooling on every architecture, and under first-token pooling on three of four (on Mistral-7B, $14.2\%$ against $20.5\%$). Absolute retrieval rates depend on whether the chosen pooling captures content.

\begin{table}[h]
\centering
\caption{Pooling sensitivity (hard CounterFact retrieval). Mean pooling is best for all architectures, and sheaf $H^0$ beats the full hidden state under every pooling choice except first-token pooling on Mistral-7B.}
\label{tab:pooling_sensitivity}
\small
\begin{tabular}{@{}lcccccc@{}}
\toprule
& \multicolumn{2}{c}{\textbf{Mean (default)}} & \multicolumn{2}{c}{\textbf{Last token}} & \multicolumn{2}{c}{\textbf{First token}} \\
\textbf{Model} & Full & $H^0$ & Full & $H^0$ & Full & $H^0$ \\
\midrule
GPT-2          & 43.1 & \textbf{85.3} & 25.5 & 74.3 & 41.8 & 42.7 \\
Phi-2          & 69.1 & \textbf{87.1} & 26.9 & 65.9 & 30.5 & 43.8 \\
Mistral-7B     & 77.7 & \textbf{87.1} & 40.4 & 72.2 & 20.5 & 14.2 \\
Llama-2-7B     & 68.2 & \textbf{80.4} & 21.4 & 47.0 & 19.6 & 23.0 \\
\bottomrule
\end{tabular}
\end{table}

\subsection{Experimental Details}
\label{app:details}

\textbf{Compute:} All computations performed on NVIDIA A100 (80GB) or RTX 4090 (24GB) GPUs. Models with 7B parameters or more are 4-bit quantised using bitsandbytes. It takes $< 1$ minute to construct the sheaf (that is, learn a restriction map and compute the Laplacian) for each model, although extracting activations will generally be much more expensive, as with any activation-based method.

\textbf{Hyperparameters:} $k = 128$, explaining $\geq 98\%$ of the total variance. All sequence positions equally weighted (mean pooling). Joint PCA over all pairs to obtain restriction map. The 20 lowest and highest eigenvectors are used as the $H^0$ and $H^1$ bases respectively. All steering vectors scaled to 30\% of the average norm of hidden states.

\textbf{Layer selection:} The relevant layer for retrieval, and all steps in constructing the sheaves, is $\ell = \lfloor 0.67L \rfloor$, that is, layer 8 for GPT-2 and layer 21 for 32-layer models. For steering, the relevant layer is $\ell = \lfloor 5L/8 \rfloor$, layer 7 for GPT-2 and layer 20 for 32-layer models. These are all reasonably high up in the network, high enough to expect meaningful semantics, but not so high as to be strongly influenced by the output projections.

\textbf{Statistical tests:} 1000 bootstrap samples to give confidence intervals. Mann-Whitney U test between groups of examples; McNemar's test where there are paired yes-no decisions. Cohen's $d$ and $h$ (for proportions) to measure magnitude of differences. Two-tailed $p$-values throughout, except for Spearman correlation and the holonomy permutation tests, for which a one-tailed exact $p$-value in the pre-specified direction is calculated.

\subsection{Practitioner Guidance}
\label{app:practitioner}

Steps to follow when using this method: (1) Create your sheaf using at least 200 paraphrase pairs from MRPC. Use PCA to create the restriction maps. (2) Steer at an $\alpha$ around 0.10--0.15 using the $H^1$ dims and $H^0$-removed. Choose whichever performs best on a validation set of size $50$. (3) If one random-steering probe shows your model is susceptible to generation collapse, then use the sheaf $H^1$ to prevent this. If your model is robust, random steering can perform similarly to sheaf steering at $\alpha \geq 0.30$.

\subsection{NeurIPS Paper Checklist}
\label{app:checklist}

\begin{enumerate}
\item \textbf{Claims} \\
\textbf{Question:} Do the main claims made in the abstract and introduction accurately reflect the paper's contributions and scope? \\
\textbf{Answer:} \answerYes{} \\
\textbf{Justification:} 
The claims within Section~\ref{sec:intro}, are validated within \S\ref{sec:discussion} and \S\ref{sec:experiments}. 

\item \textbf{Limitations} \\
\textbf{Question:} Does the paper discuss the limitations of the work performed by the authors? \\
\textbf{Answer:} \answerYes{} \\
\textbf{Justification:} We discuss our limitations within Section~\ref{sec:discussion}.

\item \textbf{Theory, Assumptions and Proofs} \\
\textbf{Question:} For each theoretical result, does the paper provide the full set of assumptions and a complete (and correct) proof? \\
\textbf{Answer:} \answerYes{} \\
\textbf{Justification:} There is a single Proposition (Proposition~\ref{prop:laplacian_reduction}), showing that the sheaf Laplacian reduces to projected within-pair covariance. The required conditions are stated in the proposition itself: the shared-projection matching-graph regime, with only one $P$ applying to all nodes and $Q_{uv} = I$ on all edges. The conditions for breakdown are similarly identified, namely allowing $P_v$ to vary over nodes with Procrustes transports. Appendix~\ref{app:cohom_validation} gives the full proof. To avoid ambiguity over whether algebraic or spectral $H^1$ is intended, see Remark~\ref{rem:h1}.

\item \textbf{Experimental Result Reproducibility} \\
\textbf{Question:} Does the paper fully disclose all the information needed to reproduce the main experimental results of the paper to the extent that it affects the main claims and/or conclusions of the paper? \\
\textbf{Answer:} \answerYes{} \\
\textbf{Justification:} Equations summarizing the full pipeline are given in \S\ref{sec:methodology}: the restriction map from joint PCA; the sheaf Laplacian as projected within-pair covariance; the smallest and largest eigenspaces as $H^0$ and $H^1$ respectively; and mean-difference steering after removing $H^0$. Hyperparameters are given in \S\ref{sec:experiments} and Appendix~\ref{app:details}: edge dimension $k = 128$; $m = 20$ for both $H^0$ and $H^1$; layer $\ell \approx 2L/3$ ($\lfloor 5L/8 \rfloor$ when steering); steering norm $\alpha = 0.3 \|\bar h\|$; MLP encoder geometry $d \to 256 \to 256 \to 64$; numbers of training steps. All data is publicly available and standard splits are used, with facts split 70/30 train/test for the comparison against LEACE. 7B+ models are run with 4-bit quantization, and the layer selection rules are stated.

\item \textbf{Open Access to Data and Code} \\
\textbf{Question:} Does the paper provide open access to the data and code, with sufficient instructions to faithfully reproduce the main experimental results, as described in supplemental material? \\
\textbf{Answer:} \answerYes{} \\
\textbf{Justification:} All datasets are public benchmarks with standard licenses, and sources are given in \S\ref{sec:experiments} and below. We open-sourced our code along with instructions to reproduce each claim, at \url{https://github.com/CWRU-AISM/gluing-lc}. We believe that the description in \S\ref{sec:methodology} and hyperparameters given in Appendix~\ref{app:details} allow the whole pipeline to be recreated from the paper alone. Results are deterministic given the published hyperparameters and seeds.

\item \textbf{Experimental Setting/Details} \\
\textbf{Question:} Does the paper specify all the training and test details (e.g., data splits, hyperparameters, how they were chosen, type of optimizer, etc.) necessary to understand the results? \\
\textbf{Answer:} \answerYes{} \\
\textbf{Justification:} For all 9 models, we provide the sizes, layers for retrieval and steering ($\ell \approx 2L/3$, or $\lfloor 5L/8 \rfloor$), $k = 128$, $m = 20$ dimensions in $H^0$ and $H^1$, restriction maps, and dataset splits and sample sizes. See Section~\ref{sec:experiments} (Setup and Datasets) and Appendix~\ref{app:details}.

\item \textbf{Experiment Statistical Significance} \\
\textbf{Question:} Does the paper report error bars suitably and correctly defined or other appropriate information about the statistical significance of the experiments? \\
\textbf{Answer:} \answerYes{} \\
\textbf{Justification:} We provide bootstrap CIs fpr all steering and retrieval estimates (see Tables~\ref{tab:causal_nine_arch}, \ref{tab:steering_four_arch} and \ref{tab:leace_held_out}). We also apply McNemar's exact test to pairs of yes-no outcomes, calculate Cohen's $d$, and apply Mann--Whitney $U$ test to unpaired samples. We correct for multiple testing using the Benjamini--Hochberg procedure, as discussed in Appendix~\ref{app:significance}. For MLST we run 10 random seeds, show the standard deviation and apply a paired $t$-test. Except for Spearman correlation and the holonomy permutation tests, all $p$-values are two-tailed.

\item \textbf{Experiments Compute Resources} \\
\textbf{Question:} For each experiment, does the paper provide sufficient information on the computer resources (type of compute workers, memory, time of execution) needed to reproduce the experiments? \\
\textbf{Answer:} \answerYes{} \\
\textbf{Justification:} The hardware, reported in Appendix~\ref{app:details}, is either an NVIDIA A100 (80GB) or an RTX 4090 (24GB). 7B+ models are run with 4-bit quantization. It takes $< 1$ minute per model to construct the sheaf, and extracting activations is the dominant cost. As described in Section~\ref{sec:experiments}, no experiment needs more than a single A100.

\item \textbf{Code of Ethics} \\
\textbf{Question:} Does the research conducted in the paper conform, in every respect, with the NeurIPS Code of Ethics? \\
\textbf{Answer:} \answerYes{} \\
\textbf{Justification:} This is purely an interpretability study, carried out on a range of publicly available pre-trained language models, evaluated on publicly available paraphrase and factual-recall benchmarks. There is therefore no issue of experiments on people or crowdsourcing, scraping of private data, or creation and dissemination of new high-risk artifacts. All relevant datasets (MRPC, PAWS, QQP, STS-B, CounterFact) are standard licensed benchmarks. We have read the NeurIPS Code of Ethics, and believe we abide by all aspects.

\item \textbf{Broader Impacts} \\
\textbf{Question:} Does the paper discuss both potential positive societal impacts and negative societal impacts of the work performed? \\
\textbf{Answer:} \answerYes{} \\
\textbf{Justification:} This paper seeks to increase the transparency of the representations learned by large language models. It could be beneficial for identifying without labels which directions in residual streams are content stable under paraphrasing and which carry surface variation, to assist red-teamers, test alignment, or diagnose model fragility (see analysis in \S\ref{sec:steering_diagnostic}). The same techniques could in principle be used to alter models' output style or stance in order to subvert safety guardrails but our intention in this study is purely as a diagnostic.

\item \textbf{Safeguards} \\
\textbf{Question:} Does the paper describe safeguards that have been put in place for responsible release of data or models that have a high risk for misuse (e.g., pretrained language models, image generators, or scraped datasets)? \\
\textbf{Answer:} \answerNA{} \\
\textbf{Justification:} No new pretrained models, generative models or scraped datasets have been created. A variety of publicly available models (GPT-2, Llama-2/3, Mistral, Qwen, phi-2, Gemma and OLMo) have been studied, against existing public benchmarks. Hence only code to reproduce our analyses is made available, not a model checkpoint or dataset.

\item \textbf{Licenses for existing assets} \\
\textbf{Question:} Are the creators or original owners of assets (e.g., code, data, models), used in the paper, properly credited and are the license and terms of use explicitly mentioned and properly respected? \\
\textbf{Answer:} \answerYes{} \\
\textbf{Justification:} All datasets and models are referenced and licenses respected: MRPC~\cite{dolan2005mrpc}, PAWS~\cite{zhang2019paws}, QQP~\cite{iyer2017qqp}, CounterFact~\cite{meng2022locating} and TruthfulQA~\cite{lin2022truthfulqa}. 

\item \textbf{New Assets} \\
\textbf{Question:} Are new assets introduced in the paper well documented and is the documentation provided alongside the assets? \\
\textbf{Answer:} \answerNA{} \\
\textbf{Justification:} There are no new datasets, pretrained models or benchmarks. We provide the analysis code and reproduction guide.

\item \textbf{Crowdsourcing and Research with Human Subjects} \\
\textbf{Question:} For crowdsourcing experiments and research with human subjects, does the paper include the full text of instructions given to participants and screenshots, if applicable, as well as details about compensation (if any)? \\
\textbf{Answer:} \answerNA{} \\
\textbf{Justification:} There is no crowdsourcing or other experimentation with human subjects. 

\item \textbf{Institutional Review Board (IRB) Approvals or Equivalent for Research with Human Subjects} \\
\textbf{Question:} Does the paper describe potential risks incurred by study participants, whether such risks were disclosed to the subjects, and whether Institutional Review Board (IRB) approvals (or an equivalent approval/review based on the requirements of your country or institution) were obtained? \\
\textbf{Answer:} \answerNA{} \\
\textbf{Justification:} No human subjects are involved; no IRB approval is required.

\item \textbf{Declaration of LLM Usage} \\
\textbf{Question:} Does the paper describe the usage of LLMs if it is an important, original, or non-standard component of the core methods in this research? Note that if the LLM is used only for writing, editing, or formatting purposes and does not impact the core methodology, scientific rigorousness, or originality of the research, declaration is not required. \\
\textbf{Answer:} \answerNA{} \\
\textbf{Justification:} The paper analyzes pretrained LLMs (GPT-2, Llama-2/3, Mistral, Qwen, phi-2, Gemma, OLMo) as objects of study; it does not use LLMs as part of its core methodology to generate science content, design the cohomological framework, derive the proposition, or run the experiments. Any LLM-assisted writing or editing is non-substantive and does not change methodology, rigor, or originality.
\end{enumerate}

\end{document}